\documentclass[a4paper,fleqn]{cas-sc}
\usepackage[numbers,sort&compress]{natbib}
\usepackage{amsmath}
\usepackage{eucal}
\usepackage{geometry}
\usepackage{graphicx}
\usepackage{subcaption}
\usepackage{algpseudocode}
\usepackage{algorithm}
\usepackage{amsfonts}
\usepackage{hyperref}
\usepackage{diagbox}
\usepackage{xcolor}
\usepackage{boondox-cal}
\usepackage{threeparttable}
\usepackage{pifont}
\usepackage{listings}
\usepackage{amsfonts,amssymb}
\usepackage{mathrsfs}
\usepackage{boondox-cal}
\usepackage{lipsum,multicol}
\usepackage{cleveref}
\crefname{figure}{Fig.}{figures}
\newcommand{\myref}[1]{Eq. (\ref{#1})}
\newcommand{\myyyref}[1]{Table \ref{#1}}
\newcommand{\myyyyyref}[2]{Algorithm \ref{#1}}
\newcommand{\doi}[1]{\href{#1}{#1}}
\definecolor{codegreen}{rgb}{0,0.6,0}
\definecolor{codegray}{rgb}{0.5,0.5,0.5}
\definecolor{codepurple}{rgb}{0.58,0,0.82}
\definecolor{backcolour}{rgb}{0.95,0.95,0.92}
\lstdefinestyle{mystyle}{
  backgroundcolor=\color{white}, commentstyle=\color{codegreen},
  basicstyle=\footnotesize\ttfamily\tiny,
  keywordstyle=\color{magenta},
  numberstyle=\tiny\color{codegray},
  stringstyle=\color{codepurple},
  basicstyle=\ttfamily\footnotesize,
  breakatwhitespace=false,
  breaklines=true,
  captionpos=b,
  keepspaces=true,
  numbers=none,
  numbersep=5pt,
  showspaces=false,
  showstringspaces=false,
  showtabs=false,
  tabsize=2
}
\begin{document}
\shorttitle{}
\shortauthors{C. He, H.M. Shi et~al.}%
\title [mode = title]{Modulated differentiable STFT and balanced spectrum metric for freight train wheelset bearing cross-machine transfer monitoring under speed fluctuations}
\author[1,2]{Chao He}[style=chinese]%
\ead[URL]{chaohe@bjtu.edu.cn}%
\author[1,2]{Hongmei Shi}[style=chinese]
\ead[URL]{hmshi@bjtu.edu.cn}%
\cormark[1]
\cortext[cor1]{Corresponding author.}
\author[1,2]{Ruixin Li}[style=chinese]
\ead[URL]{20221357@bjtu.edu.cn}%
\author[1,2]{Jianbo Li}[style=chinese]
\ead[URL]{jbli@bjtu.edu.cn}%
\author[1,2]{ZuJun Yu}[style=chinese]
\ead[URL]{zjyu@bjtu.edu.cn}%
\address[1]{State Key Laboratory of Advanced Rail Autonomous Operation, Beijing Jiaotong University, Beijing 100044, China}
\address[2]{School of Mechanical, Electronic and Control Engineering, Beijing Jiaotong University, Beijing 100044, China}

\begin{abstract}
The service conditions of wheelset bearings has a direct impact on the safe operation of railway heavy haul freight trains as the key components. However, speed fluctuation of the trains and few fault samples are the two main problems that restrict the accuracy of bearing fault diagnosis. Therefore, a cross-machine transfer diagnosis (pyDSN) method coupled with interpretable modulated differentiable short-time fourier transform (STFT) and physics-informed balanced spectrum quality metric is proposed to learn domain-invariant and discriminative features under time-varying speeds. Firstly, due to insufficiency in extracting extract frequency components of time-varying speed signals using fixed windows, a modulated differentiable STFT (MDSTFT) that is interpretable with STFT-informed theoretical support, is proposed to extract the robust time-frequency spectrum (TFS). During training process, multiple windows with different lengths dynamically change. Also, in addition to the classification metric and domain discrepancy metric, we creatively introduce a third kind of metric, referred to as the physics-informed metric, to enhance transferable TFS. A physics-informed balanced spectrum quality (BSQ) regularization loss is devised to guide an optimization direction for MDSTFT and model. With it, not only can model acquire high-quality TFS, but also a physics-restricted domain adaptation network can be also acquired, making it learn real-world physics knowledge, ultimately diminish the domain discrepancy across different datasets. The experiment is conducted in the scenario of migrating from the laboratory datasets to the freight train dataset, indicating that the hybrid-driven pyDSN outperforms existing methods and has practical value.
\end{abstract}

\begin{highlights}
\item A pyDSN is proposed for heavy haul freight train wheelset bearing transfer diagnosis.
\item pyDSN can tackle cross-machine transfer diagnosis under speed fluctuations.
\item The modulation differentiable STFT incorporated a mask modulation mechanism.
\item Balanced spectrum is devised to evaluate the quality of spectrograms.
\item The interpretability of MDSTFT is elaborated from both quantitative and qualitative perspectives.
\end{highlights}


\maketitle
\section{Introduction}\label{section:01}
Parallel to the ongoing increase in load weight and operational mileage, heavy haul freight trains are experiencing a decline in service condition. Bearing failures can compromise the dynamic performance of train operations, potentially leading to severe issues such as axle overheating and axle breakage, which may result in significant operational safety incidents. Consequently, monitoring the health status of bearings and diagnosing faults have followed with interest for railway departments and researchers.

Specially focused on wheelset bearings, schemes based on signal processing or traditional machine learning have been investigated. Mykhalkiv et al. \cite{h45}  presented a signal processing algorithm based on minimum entropy deconvolution and square envelope spectra for axle-box bearing faults. Wang et al. \cite{h46} designed a feature engineering algorithm for freight train air brakes. Si et al. \cite{h47} verified the effectiveness of utilizing temperature data for monitoring bearing faults. Wu et al. \cite{h69} applied a two-stage strategy that combined autocorrelation and an improved flow Gaussian mixture. These plans all rely on expert experience. Nowadays, artificial intelligence (AI)-based intelligent diagnosis, with end-to-end feature representation capabilities, has drawn considerable acclaim in the current era. Liu et al. \cite{h44} contributed to electronically controlled pneumatic utilizing feature fusion and ensemble learning techniques. Li et al. \cite{h48} proposed a transfer learning wheelset bearing fault diagnosis algorithm based on fast fourier transform (FFT) and fine-tuned convolutional neural networks (CNN) for heavy haul freight train, but it required target domain annotated dataset that come from the same equipment at constant speed. In summary, owing to the sparse discourse, it is currently quite immature within the realm of AI-driven heavy haul freight train wheel bearing monitoring. For another matter, the aforementioned techniques failed to concurrently address the principal pain points from two fronts in heavy haul freight train wheel diagnosis as follows.
\begin{enumerate}[\dag]
\item For the first challenge, in the process of actual operation, the heavy haul freight trains inevitably experience acceleration and deceleration process, generating copious non-stationary signals possessing fault information that cannot be found at constant speed. It impedes the efficacy of the algorithms predicated on constant speed \cite{h101}.
\item Meanwhile, it is time-consuming and labor-intensive to obtain sufficient annotated fault data, which hampers the practicality of supervised learning-based intelligent fault diagnosis schemes.
\end{enumerate}

In summary, the issue can be defined as the unsupervised cross-machine transfer learning diagnosis for heavy haul freight train wheel bearings under conditions of small samples and fluctuating speeds, which represents a novel issue, an uncharted challenge that has not been previously reported.

In response to the first challenge of fluctuating speeds, extensive advancements \cite{h101,h832,h72,h200} have been made. The traditional signal processing approaches covers order tracking \cite{h102}, cyclic spectrum correlation, generalized demodulation \cite{h104}, and time-frequency analysis \cite{h105}. Although aforementioned methods exhibit an unambiguous interpretability ascribed to rigorous derivation, the dependence on expert experience hinders the execution of automated diagnosis under time-varying speeds. As the second dimension, the AI-assisted time-varying speed intelligent fault diagnosis can enable the end-to-end mode. Li et al. \cite{h11} and Shao et al. \cite{h111} introduced machine learning to identify small infrared thermal images of various faults at variable speeds. Xu et al. \cite{h12} encoded signals under linear varying speeds into images utilizing STFT, and subsequently input them to capsule network and Transformer to examine the diagnosis capacity under small samples. Rao et al. \cite{h13} argued that speed changes can engender imbalanced fault information, and thus suggested a speed normalization module to weaken this impact. Meanwhile, they further devised a plug and play branch named speed adaptive gate\cite{h14}.
In addition, knowledge decoupling framework \cite{h16}, PeriodNet \cite{h15}, anti-symmetric Laplace-Stacked Autoencoder \cite{h17}, alternative kernel convolutional networks \cite{h18}, stack denoising autoencoders and gated recurrent unit \cite{h19}, long short-term memory \cite{h20}, multi-branch redundant adversarial net \cite{h21}, restricted sparse networks \cite{h211}, deep nonlinear order-cyclic convolutional network \cite{h71}, graph attention neural networks \cite{h22}, attention mechanism \cite{h38, h834} and generation network \cite{h836} have also achieved success in diverse variable speed research.
While the methodologies outlined alleviate the first issue, they fails to deal with the cross-domain challenge. It is because that obtaining a multitude of heavy haul freight train data is impractical. With a plethora of annotated data from laboratory equipment available, cross-machine transfer diagnosis \cite{h10} within fluctuating speed scenarios is emerging as a promising idea for the second challenge that the targeted device poses challenge of limited samples.

Despite few research dedicated to cross-machine transfer diagnosis within fluctuating speed scenarios, transfer-learning diagnosis within the same machine has yielded noteworthy outcomes \cite{h74, h833,h835}.
For domain discrepancy metric, Liang et al. \cite{h24} leveraged continuous wavelet transform (CWT), deformable convolution, and local maximum mean discrepancy (LMMD) to propose a diagnosis method for the same machine from single source domain variable speed data to multiple target domains. Subsequently, their investigation \cite{h25} primarily concentrated on two tasks of constant speed to variable speeds, and variable speeds to variable speeds for the same machine.
Cao et al. \cite{h30} primarily targeted the research of constant speed to variable speeds and suggested Cauchy kernel induced maximum mean dispersion (CK-MMD) to address the domain shift. Si et al. \cite{h28} designed a multi-order moment matching loss. Zhao et al. \cite{h6} put forward a graph convolutional neural network based on dynamic multi-kernel maximum mean discrepancy (DMK-MMD) for aircraft engine diagnosis under different loads, which utilized the order spectrum analysis to obtain the order spectrum signal.

Concerning an alternate category of approaches that involve data preprocessing and model design, Shi et al. \cite{h27} proposed a reliable feature-assisted contrastive generalization net, which efficiently achieved the migration of variable speed data from multiple source domains to single target domain. Zhou et al. \cite{h29} modified the self-attention module to depthwise separable convolution and utilized comparative regularization to assist the improved Transformer to extract transferable features for low and high speeds both, enabling the detection of medium-speed faults. Subsequently, Gao et al. \cite{h26} established a dual correlation model based on graph neural networks to facilitate the transferring from low-speed regions to high-speed regions, verifying the reliability of the proposed algorithm. Xu et al. \cite{h32} came up with an algorithm formulated based on synchronous compressed wavelet transform-time feature order spectrum and multi-scale domain adaptation networks (DAN) to substantiate its capability in detecting faults across all speeds and between acceleration and deceleration of the same machine. Lu et al. \cite{h33} devised a spectrum alignment and DAN to cope with unbalanced issue at a wide range of speed variation conditions. Pang et al. \cite{h42} introduced a time-frequency supervised contrastive learning framework for cross-speed fault diagnosis of bearings, which achieves the transfer diagnosis from constant speed to acceleration or deceleration in various bearing health conditions. Luo et al. \cite{h68} introduced a meta-learning approach that employs an elastic prototypical network (EProtoNet) for few-shot fault transfer diagnosis in scenarios ranging from stable to unstable speeds on the same device. However, the issue of cross-machine transfer diagnosis in scenarios on speed fluctuations has been overlooked. Concurrently, the aforementioned reviewed cross-domain algorithms are purely data-driven, which critically depend upon the quality and quantity of data. They disregard the significant role of interpretable signal driving and physical constraints on in fostering cross-domain transfer diagnosis.

Notably, time-frequency representation using STFT serves as an effective way to process non-stationary signals under time-varying speeds. However, when the speed varies, the periodicity and amplitude of the impact caused by the fault will alter, and the fault characteristic frequency will also evolve over time, producing more complex non-stationary signals. When STFT is adopted, if the window length is too long, it results in higher frequency resolution, while time resolution will decrease. Therefore, for data under speed fluctuations, while differentiable STFT can automatically search for the optimal window length \cite{h34,h35,h66}, utilizing a fixed window length still remains ineffective and tricky.

In conclusion, the current study does possess certain limitations.
\begin{enumerate}[\dag]
\item To the best of our knowledge, in the realm of heavy haul freight trains, there remains few research on cross-machine bearing diagnosis under variable speeds, and it has not been documented and studied in detail yet. Also, the reported cross-domain algorithms disregard the beneficial impact of interpretable signal driving and physical constraints on transfer diagnosis.
\item Current cross-domain investigations concentrate on different speeds and loads from the same device. Owing to the inability to collect sufficient labels from heavy haul freight train wheels, cross-domain algorithms for the same machine are ill-equipped to address this problem of cross-machine diagnosis under time-varying speeds.
\item The existing approaches to cross-domain variable speeds follow the two-stage strategy, namely involving preliminary preprocessing followed by domain adaptation. The effectiveness hinges extensively upon the quality of preprocessing, while preprocessing relies on manual experience. It not only results in cumbersome parameter selection, but also hinders automated end-to-end diagnosis.
\item While differentiable STFT has shown promise in fault diagnosis, all signal samples adopt a fixed windows length, which is not conducive to capturing distinctive fault information contained in variable speeds, and the cost-sensitive loss does not contemplate the inherently physical information expression of STFT.
\end{enumerate}

In accordance with our prior work---IDSN \cite{h10}, interpretable differentiable STFT is considered as "prior knowledge", DAN counts as "data-driven techniques". In this paper, Balanced Spectrum Quality metric is "physics-informed expression". A knowledge-data-physics-informed one-stage triple-driven model, termed as pyDSN, is established. pyDSN can alleviate both severe dependence on homemade parameters and quality and quantity of datasets. With the support of domain discrepancy metric, pyDSN can also achieve bearing monitoring from laboratory simulation data to real-world heavy haul freight train wheel bearings.

In comparison with IDSN (DSTFT) \cite{h10}, pyDSN (MDSTFT) has been substantively improved. For data under speed fluctuations, while IDSN can automatically search for the optimal window length, utilizing a fixed window length still remains ineffective and tricky. Because variable speed data contains some fault information that does not exist absent from constant one, a differentiable variable window length STFT algorithm is proposed. The time-varying fixed windows in IDSN are modified to time-varying and variable windows to facilitate the fault information expression. In addition, IDSN does not contemplate the inherently physical information expression of STFT.

The physical information in this paper, specifically refers to the physics-informed constraints, which serve as a regularization term to alleviate overfitting and enhance the robustness of model. The physics-informed loss integrated with physical prior knowledge can constrain the optimization path, designing task-specific loss functions. The weights of each loss item are balanced, ensuring a lucid fusion of physical prior and neural networks, thereby enhancing interpretability \cite{h75,h76}. Liao et al. \cite{h77} introduced kurtosis and $G - {l_1}/{l_2}$ norm for blind deconvolution networks. Yan et al. \cite{h78} proposed a physics-guided loss function for modeling healthy degradation; Russell et al. \cite{h79} incorporated frequency content into Auto-encoder. Chen et al. \cite{h80} represented physical degradation characteristics as loss function constraints to guide network training. Freeman et al. \cite{h81} constructed a physics-informed loss to bridge the gap between fault features and the environment. Xu et al. \cite{h82} designed a physics-aware loss for building damage assessment; Zhang et al. \cite{h83} developed a physical loss function for structural health monitoring to evaluate the difference between model output and finite element output. Nevertheless, for the arduous task of cross-machine transfer diagnosis under speed fluctuations, there remains still a lack of a physics-informed loss capable of achieving high-performance transfer diagnosis.

To bridge this gap, rather than mere classification and measurement losses as in IDSN, the physics-informed constraint, referred to as BSQ, has been enforced in the training process of pyDSN, assessing the time-frequency resolution of TFS. The physical loss is incorporated into the training procedure to extract more generalized feature representations. We argue that the infliction of physical constraints is essential to acquire transferable and discriminative features. This study proposes a taxonomy with two dimensions to assess the TFS quality generated by STFT:
\begin{enumerate}[\dag]
\item One type is the evaluation indicator, which evaluates the quality of the generated STFT through a certain calculation method, such as Intelligent spectrogram \cite{h39}, Rényi entropy \cite{h41}, and so forth.
\item Another type is the differentiable loss index, encompassing all attributes of the evaluation indicator while possessing certain unique features.
\end{enumerate}

On the one hand, due to its capacity to backpropagate and adaptively modify pivotal parameters in MDSTFT, it can steer MDSTFT towards generating the superior STFT spectrogram; On the other hand, the differentiable loss index unifies the standard of distinct evaluation indicators, circumventing the scenario where one evaluation indicator ranks high and another evaluation indicator rates poorly, achieving trade-off among various evaluation indicators. In differentiable loss indexes, the trusted metrics based on kurtosis and entropy are two commonly employed methodologies, but these metrics are usually ineffective for pulse signal components of the rotating mechanical equipment \cite{h40}. Consequently, balanced spectrum quality loss is devised to evaluate the physical properties of time-frequency spectrograms and ultimately serve as a physical constraint to augment the extrapolation capability of domain adaptation networks.

Lastly, in the realm of intelligent fault diagnosis, the cross-machine diagnosis from time-varying speeds to time-varying speeds remains an innovative topic that has yet to be explored. PyDSN is the first work, as a starting point food for thought.

The key findings of this investigation encompass the following four aspects.
\begin{enumerate}[\dag]
\item A physics-informed cross-machine transfer method, named as pyDSN, is proposed for heavy haul freight train wheels to address fault detection under time-varying speed conditions. It features two state-of-the-art modules: modulated differentiable STFT (MDSTFT) and balanced spectrum quality loss (BSQ).
\item Distinct from employing a fixed window length, MDSTFT with time-varying window is devised, which adaptively adopts different time-varying window lengths to facilitate the extraction of fault details contained at various rotation speeds. The windows are adaptively adjusted and the length of each is different.
\item As a third kind of metric to capture transferable features, BSQ is devised that can comprehensively embody the physical information of time-frequency spectrum and serve as a regularization. It can steer the differentiable time-varying STFT algorithm towards efficient TFS and also regulate DAN within physical boundaries to learn real-world physics knowledge.
\item Comprehensive analysis illustrates pyDSN demonstrates commendable functionality in time-varying operating conditions, substantiating the reliability and superior performance with the proposed methodology. This is the first successful attempt for cross-machine transfer diagnosis under speed fluctuation.
\end{enumerate}
\vspace{-0.1cm}

Subsequent sections are organized into the following structure. We start with defining the main problem and reviewing the prior theories. The various components of pyDSN are elaborately outlined in Section \ref{section:03}. Section \ref{section:04} incorporates multiple empirical trials and assessments to emphasise the inherent advancements. Finally, Section \ref{section:05} conducts a thorough interpretability analysis. It concludes in Section \ref{section:06}.
\section{Preliminaries}\label{section:02}
\subsection{Problem formulation}
The present issue may be characterized as the transfer diagnosis problem for bearings on heavy haul freight train wheels, in scenarios involving small samples and speed fluctuations.

\begin{enumerate}[\dag]
\item Owing to the complexity of data and models, "small sample" is difficult to delineate consistently and precisely. Typically, in machine learning, it can lead to over-fitting. In scenarios where the training data is scarce, the learned feature representation is limited and only fits the training data well, resulting in superior training accuracy yet diminished test precision.
     According to Ref. \cite{h12}, the size of training set is 100K (K is the fault category), and thus the task belongs to small sample issue.
\item The concept of "speed fluctuation" can be summarized as the collected signals not being at the constant speed condition, but at a linearly varying speed, because heavy haul freight trains inevitably encounter the reality of acceleration or deceleration.
\item Cross-machine transfer diagnosis stems from the fact that the source domain has numerous labels, whereas the target domain lacks labels. This is because a large amount of laboratory data can be available, but there is less labelled data on real-world heavy haul freight train wheel bearings.
\end{enumerate}
\vspace{-0.1cm}

The source domain and target domain can be respectively rendered as ${D_s}$:${X^s_m} = {\{ \left. {(x_i^s,y_i^s)} \right|i = 1,2,3, \ldots ,{n_s}\}}$ and ${D_t}$:${X^t _m}={\{ \left. {(x_j^t,y_j^t)} \right|j = 1,2,3, \ldots ,{n_s}\}}$. Among them, $m$ is the number of samples, $i, j $ denote the number of sample categories, $m \le 20 $. $x$ is the sample, and $y $ is label. $ s $ represents the source domain, and $t $ represents the target domain. The variable speed dataset amassed from the laboratory in the source domain has a data distribution $P_s$. The target domain is the real-world heavy haul freight train wheel dataset, with a data distribution $P_ t$.

The task of the paper objective lies in minimizing the $ \Im=<P_ s. P_ t>$  at first during the data input stage. Simultaneously, in the classification stage, further diminish the size of $ \Im $ in Reproducing kernel Hilbert Space (RKHS). In conclusion, the risk loss can be mitigated via two strategies, namely physical-informed data augmentation and minimized domain discrepancy. In order to rectify these concerns, a physical-informed modulated differentiable STFT-based cross-machine diagnosis for heavy wagons is proposed.
\subsection{Related works about differentiable STFT}
Studies into DSTFT have predominantly employed two distinct methodologies\cite{h10,h35,h37}.
\subsubsection{Scheme One}
In the seminal work pertaining to DSTFT applied to cross-machine transfer diagnosis, the differentiable coefficient ($ \theta $) is introduced; subsequently, the window length and window function are adapted via $ \theta $, making the time-frequency resolution (${F_W}[t,f]$) trade-off as shown in \myref{eq1}.
\begin{flalign}\label{eq1}
&{F_W}[t,f] = \sum\limits_{n = 1}^{N - 1} {x[t + n]{W_{m,\theta }}[n]{e^{ - j\frac{{2\pi }}{N}fn}}}&
\end{flalign}
where $t$ is the sample position, $n$ is window length, $j$ is the unit imaginary number, $N$ is the support size.

Considering Kaiser window as an example, the differentiable window can be expressed as:
\begin{flalign}\label{eq2}
&\left\{ {\begin{array}{*{20}{c}}
{n = \gamma \theta }\\
\begin{array}{l}
W_{m,\theta }^{\rm{Kaiser}}[n] = \frac{{{I_0}\left[ {\beta \sqrt {1 - {{(1 - \frac{{2m}}{{n - 1}})}^2}} } \right]}}{{{I_0}(\beta )}}\\
 = \frac{{{I_0}\left[ {\beta \sqrt {1 - {{(1 - \frac{{2m}}{{\gamma \theta  - 1}})}^2}} } \right]}}{{{I_0}(\beta )}}
\end{array}
\end{array}} \right.&
\end{flalign}
where the hop/window ratio is $\gamma$, $I_0$ is the first type of modified Bessel function of zero order, and $\beta$ is any non negative real number used to adjust the shape of the Kaiser window.
\subsubsection{Scheme Two}
The initial step involves dividing signals ($x$) into sample slices ($X$) based on hop size and slice length. The time dimension proxy ($\varpi$) of the window function ($W_{N}^{\rm{Kaiser}}[\varpi]$) is differentiable. $\varpi$ as stipulated in \myref{eq2} is differentiable, but the first idea is that $n$ is differentiable, which is the main difference between the two. The obtained time-frequency spectrum is shown in \myref{eq3}:
\begin{flalign}\label{eq3}
&{{\text{F}}_{\text{W}}}{\left[ {t,f} \right]^\prime }{\text{ = DFFT(X}} \otimes W_{N}^{{\text{Kaiser}}}[\varpi]{\text{)}}&
\end{flalign}
Among them, DFFT is the discrete Fourier transform, and $\otimes$ is the element by element-wise product.

However, it is noteworthy that both the aforementioned methods utilize the same window length.
\section{The proposed method: pyDSN}\label{section:03}
\subsection{The Overview of pyDSN}\label{section:030}
The core idea lies in facilitating extraction of domain-invariant and discriminative features. The prior research predominantly achieve these objectives by meticulously designing backbones, decreasing domain metric discrepancy, or implementing sophisticated training methodologies. Nevertheless, this paper contends that the aforementioned tasks can be accomplished through prior-empowered of Signal processing knowledge, obviating intricate computations and architectural design. Given that the hyper-parameters of signal processing algorithms are generally determined by human experts, so a modulated differentiable STFT algorithm is proposed.
\begin{figure}[t]
  \center
  \includegraphics[scale=0.62]{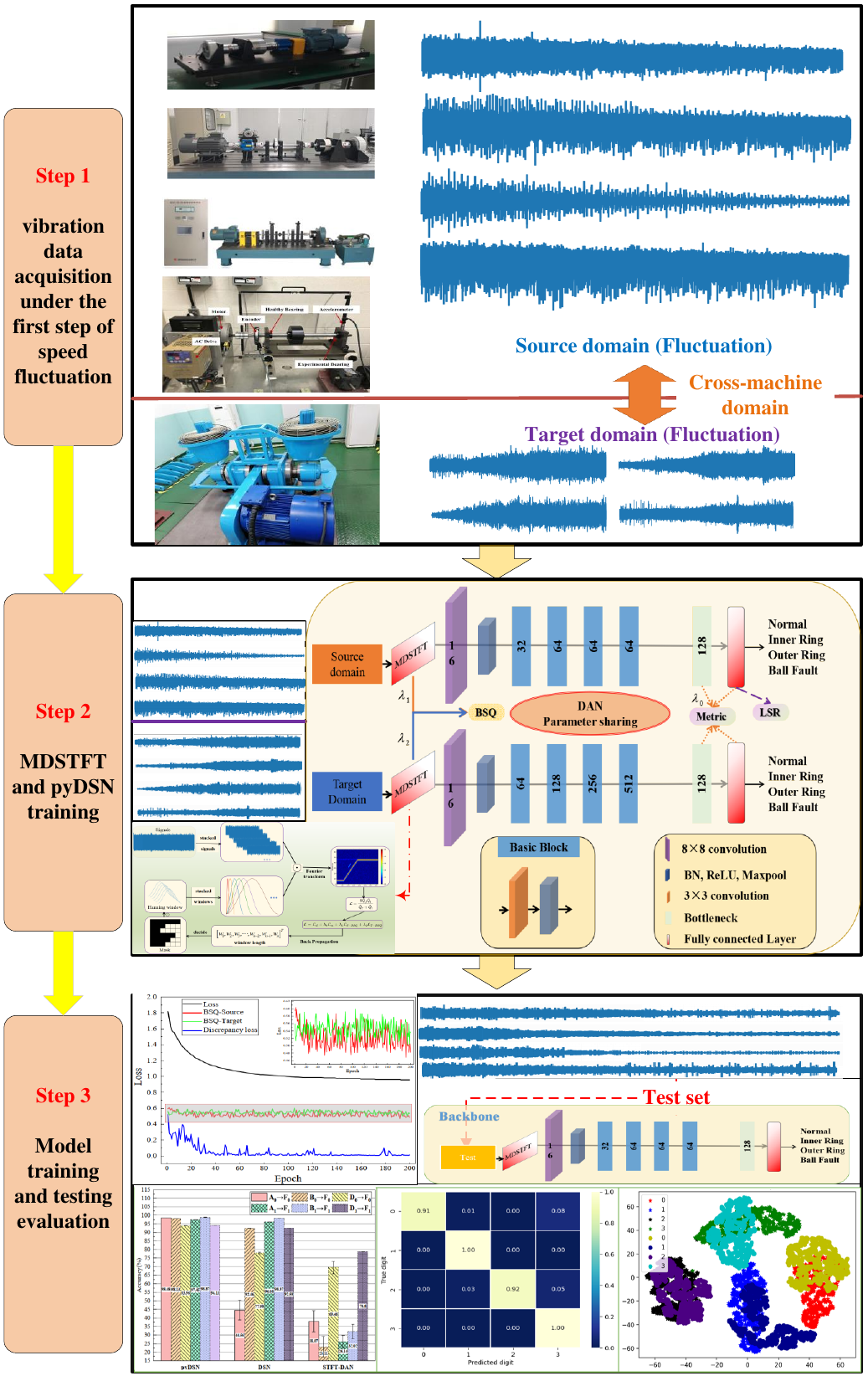}
  \caption{The overview of pyDSN.}
  \label{03}
\end{figure}

This paper constitutes the first exploration of the cross-machine transfer diagnosis scenario on speed fluctuation. The pseudo-code is as shown in Algorithm \ref{alg:01}. The systematic diagnostic procedure encompasses the following fivefold fundamental stages.
\begin{enumerate}[(1)]
\item Six vibration signals subjected to speed fluctuations are collected, encompassing one from University of Ottawa dataset, three from independently gathered bearing datasets, and two from gearbox datasets, also including a real-world heavy haul freight train wheelset bearing dataset. With the exception of the latter, the preceding five datasets serve as the source domain with an extensive volume of labels; Real-world dataset form the target domain.
\item The Source domain and target domain data are separately input into the proposed Modulated Differentiable STFT (MDSTFT), and output the STFT spectrogram. As the source and target domains originate from diverse mechanical devices, in order to extract domain invariant and discriminative features, MDSTFT modules do not share parameters.
\item The spectrogram obtained from MDSTFT is input into the unsupervised balanced spectrum quality (BSQ) loss. The lower the value of BSQ, the more concentrated energy of the spectrogram and the amplified distinctiveness of features. BSQ serves to direct the generation of spectrograms and imposes constraints on DAN. The source domain and target domain have different regularization coefficients, as shown in \myref{eq24}, where $\mathbcal{L}$ is the total loss function, and ${\lambda _0},{\lambda _1},{\lambda _2}$ are regularization coefficients, where ${\lambda _0}=\frac{{ - 4}}{{\sqrt {\frac{{epoch}}{{\max \_epoch}}}  + 1}} + 4,{\lambda _1}=1.0,{\lambda _2}=0.01$.
\item The captured spectrogram is fed into DAN and further reduce the distribution shifted between the two domains by combining label smoothing classification loss and domain metric discrepancy.
\item In the testing scenario, bearing datasets from heavy haul freight train wheelsets are fed into the trained model to test the performance.
\end{enumerate}

\begin{figure*}[h]
\begin{flalign}\label{eq24}
{\mathbcal{L} = \underbrace {{\mathbcal{L}_{cl}}}_{{\rm{Classification}}{\kern 1pt} {\rm{Loss}}} + {\lambda _0}\underbrace {{\mathbcal{L}_m}}_{{\rm{Domain}}{\kern 1pt} {\rm{Metric}}{\kern 1pt} {\rm{Loss}}} + {\lambda _1}\underbrace {{\mathbcal{L}_{S - BSQ}}}_{{\rm{Source}}{\kern 1pt} {\rm{domain}}{\kern 1pt} {\rm{Balanced}}{\kern 1pt} {\rm{Spectrum}}{\kern 1pt} {\rm{Quality}}{\kern 1pt} {\rm{Loss}}} + {\lambda _2}\underbrace {{\mathbcal{L}_{T - BSQ}}}_{{\rm{Target}}{\kern 1pt} {\rm{domain}}{\kern 1pt} {\rm{Balanced}}{\kern 1pt} {\rm{Spectrum}}{\kern 1pt} {\rm{Quality}}{\kern 1pt} {\rm{Loss}}}}
\end{flalign}
\end{figure*}

\begin{flalign}\label{eq66}
\mathbcal{L}_{m} = \left\| {\frac{1}{{{n_s}}}\mathop \sum \limits_{i,j = 1}^{{n_s}} {\phi _1}\left( {x_i^s,x_j^s} \right){\phi _2}\left( {z_i^{s},z_j^{s}} \right) - \frac{1}{{{n_t}}}\mathop \sum \limits_{i,j = 1}^{{n_s}} {\phi _1}\left( {x_i^t,x_j^t} \right){\phi _2}\left( {z_i^t,z_j^t} \right)} \right\|_\mathbcal{H}^2
\end{flalign}
where $n$ is the number of samples, $\phi$ is the kernel function, $x$ is feature, $z$ is the output of the fully connected layer.

A physics-informed pyDSN is composed of MDSTFT and BSQ loss. The overall diagnostic process is depicted in \cref{03}.
\subsection{Modulated Differentiable STFT (MDSTFT)}\label{section:031}
Given the inherent non-differentiability of vanilla STFT, an alternative approach known as DSTFT is proposed, which advocates for an adaptive adjustment of the window length according to data via gradient descent. Nevertheless, it should be noted that DSTFT is dependent on steady-state fault signals, falling short when dealing with variable speed signals.

In order to further amplify the capability of DSTFT to extract fault information, a signal modulation mechanism is proposed. With it, MDSTFT can not only perceive the most suitable window length, deriving from the probabilistic distribution of the data, but also modulate window lengths of different window functions to extract concealed variable speed fault details. In extreme cases, the window length can be infinitely akin to or equivalent to zero, indicating that the signal slice in extracting fault information will be substantially diminished or have no benefit. Therefore, this modulation scheme endows MDSTFT with flexibility in selecting the appropriate windows for extracting unique fault information within speed fluctuations.

If MDSTFT is implemented, for a certain signal slice, the window function is not a function of window length ($n$), but rather a function of  temporal dimension ($m$). Given frame ($T$), signal length ($L$), window length ($N$), hop size ($t$), the required number of frames is:
\begin{flalign}\label{eq4}
&n_T=\left\lfloor {1{\text{ + }}\frac{{L'{\text{ - N - 1}}}}{S}} \right\rfloor &
\end{flalign}
where $n_T$ is the required number of frames and $L'$ is $L$ with padding zero.

The given sample is sliced according to the time dimension, as shown in \myref{eq5}.
\begin{flalign}\label{eq5}
&\begin{array}{l}
X{\rm{  =  }}{\left[ {{X_1},{X_2}, \ldots ,{X_{{n_T}}}} \right]^T}\\
 = \left[ {\begin{array}{*{20}{c}}
{{x_1}}&{{x_2}}& \cdots &{{x_N}}\\
{{x_{1 + t}}}&{{x_{2 + t}}}& \cdots &{{x_{N + t}}}\\
 \vdots & \vdots & \vdots & \vdots \\
{{x_{1 + ({n_T} - 1)t}}}&{{x_{2 + ({n_T} - 1)t}}}& \cdots &{{x_{N + ({n_T} - 1)t}}}
\end{array}} \right]
\end{array}&
\end{flalign}

Scheme One utilizes the loop sliding window operation, which can not implement back-propagation when adopts diverse length windows. So, Scheme Two is adopted. The time proxy $\vec \varpi$ is seen as a one-dimensional vector, according to the number of frames ($n_T$), rather than a certain value in DSTFT, and $B_t$ is a sequence of equal difference vectors with a tolerance of one on the basis of support size, named the resampled window time.
\begin{flalign}\label{eq6}
&{\vec B_t}  =  {\left[ {\begin{array}{*{20}{c}}
0&1&2&\cdots &N\\
 \vdots & \vdots &\vdots & \ddots& \vdots\\
0&1&2&\cdots &N
\end{array}} \right]_{{n_T} \times N}}&
\end{flalign}
\begin{flalign}\label{eq7}
&\vec \varpi {\rm{ \in }}{\left[
{{N_1},}{{N_2},} \cdots, {{N_i}}
\right]_{{n_T} \times 1}},(N_i \le N)&
\end{flalign}

STFT for $x$ with the length of $L$ is defined as:
\begin{flalign}\label{eq8}
&{F\left[ {t,f} \right] = \sum\limits_{n = 0}^{N - 1} {x[t + n]W[n]{e^{ - j\frac{{2\pi }}{N}fn}}} }&
\end{flalign}
where $f$ is FFT frequency bin.

Vanilla DSTFT regards the window function as function of time proxy, defined as:
\begin{flalign}\label{eq9}
&{{{\hat F}_\varpi }\left[ {t,f} \right] = \sum\limits_{n = 0}^{N - 1} {x[\varpi ]W[\varpi ,{B_t}]{e^{ - j\frac{{2\pi }}{N}f\varpi }}} }&
\end{flalign}

In order to achieve differentiable and diverse window lengths, a modulation mechanism is introduced, which is ingeniously implemented through a mask matrix. The mask matrix is composed of continuous value one and $\infty$. The resampled time and time proxy are compared element by element, as illustrated in \myref{eq11}. In this way, it is feasible to adaptively mask the useless resampled time upon the learnable time proxy, so that the window function is incapable of identifying this segment, where the value of the corresponding position of the window function is $0$. Hence, a corresponding alteration and learnable ($\Delta _{m,i}$) can be achieved under varying and learnable window lengths (${{\vec \varpi }_i}$).
\begin{flalign}\label{eq11}
&{\Delta _{m,i}} = \left\{ {\begin{array}{*{20}{c}}
1,&{{{\vec B}_{t,i}} \le {{\vec \varpi }_i}}\\
\infty, &{{{\vec B}_{t,i}} > {{\vec \varpi }_i}}
\end{array}} \right.&
\end{flalign}

For example, assuming $N=5$, $n_T=3$, then ${\vec B_t} = \left[ {\begin{array}{*{20}{c}}
0&1&2&3&4
\\
0&1&2&3&4\\
0&1&2&3&4
\end{array}} \right]$ and if the learned window length is ${\vec \varpi=[5,3,1]^T}$, so ${\Delta _M}=\left[ {\begin{array}{*{20}{c}}
0&1&2&3&4\\
0&1&2&\infty &\infty\\
0&\infty &\infty &\infty &\infty
\end{array}} \right]$.

where $\mathop {\lim }\limits_{i \to \infty } W[{\Delta _M},\vec \varpi ,{\vec B_t}] = 0$.

The mask matrix is utilized to derive a modulated window function. Herein, the original form of the Kaiser window as an example:
\begin{flalign}\label{eq12}
&W^{{\rm{Kaiser}}}[n] = \frac{{{I_0}\left[ {\beta \sqrt {1 - {{(1 - \frac{{2n}}{{N - 1}})}^2}} } \right]}}{{{I_0}(\beta )}}&
\end{flalign}

The Kaiser window in DSTFT can be represented as,
\begin{flalign}\label{eq13}
&W_\varpi^{{\rm{Kaiser}}}[\varpi ,{B_t}] = \frac{{{I_0}\left[ {\beta \sqrt {1 - {{(1 - \frac{{2{B_t}}}{{\varpi  - 1}})}^2}} } \right]}}{{{I_0}(\beta )}}&
\end{flalign}

In stark contrast to the aforementioned, the modulated Kaiser window can be obtained through the mask matrix:
\begin{flalign}\label{eq14}
&W_{\vec\varpi}^{{\rm{Kaiser}}}[{\Delta _m},\vec\varpi ,{{\vec B}_t}] = \frac{{{I_0}\left[ {\beta \sqrt {1 - {{(1 - \frac{{2{\Delta _m}{{\vec B}_t}}}{{\vec \varpi  - 1}})}^2}} } \right]}}{{{I_0}(\beta )}}&
\end{flalign}

By employing \myref{eq5} and \myref{eq14}, MDSTFT is ultimately derived as displayed in \myref{eq10}.
\begin{flalign}\label{eq10}
&{{{\tilde F}_{\vec \varpi }}\left[ {t,f} \right] = \sum\limits_{n = 0}^{N - 1} {x[\vec \varpi ]W[{\Delta _M},\vec \varpi ,{{\vec B_t}}]{e^{ - j\frac{{2\pi }}{N}f\vec \varpi }}} }&
\end{flalign}

Comparing \myref{eq10} and \myref{eq9}, numerous differences can readily be discerned: Primarily, the window length is esteemed as a collection of parameters rather than a single variable, achieving adaptive adjustment of diverse window lengths. Secondly, with strictly following window function calculation in the STFT process, a modulation mechanism based on mask matrix is introduced, which not only upholds the differentiability of the window function, but also standardizes the calculation procedure, satisfying the different window functions corresponding to different lengths.
\begin{figure*}[t]
  \center
  \includegraphics[scale=0.4]{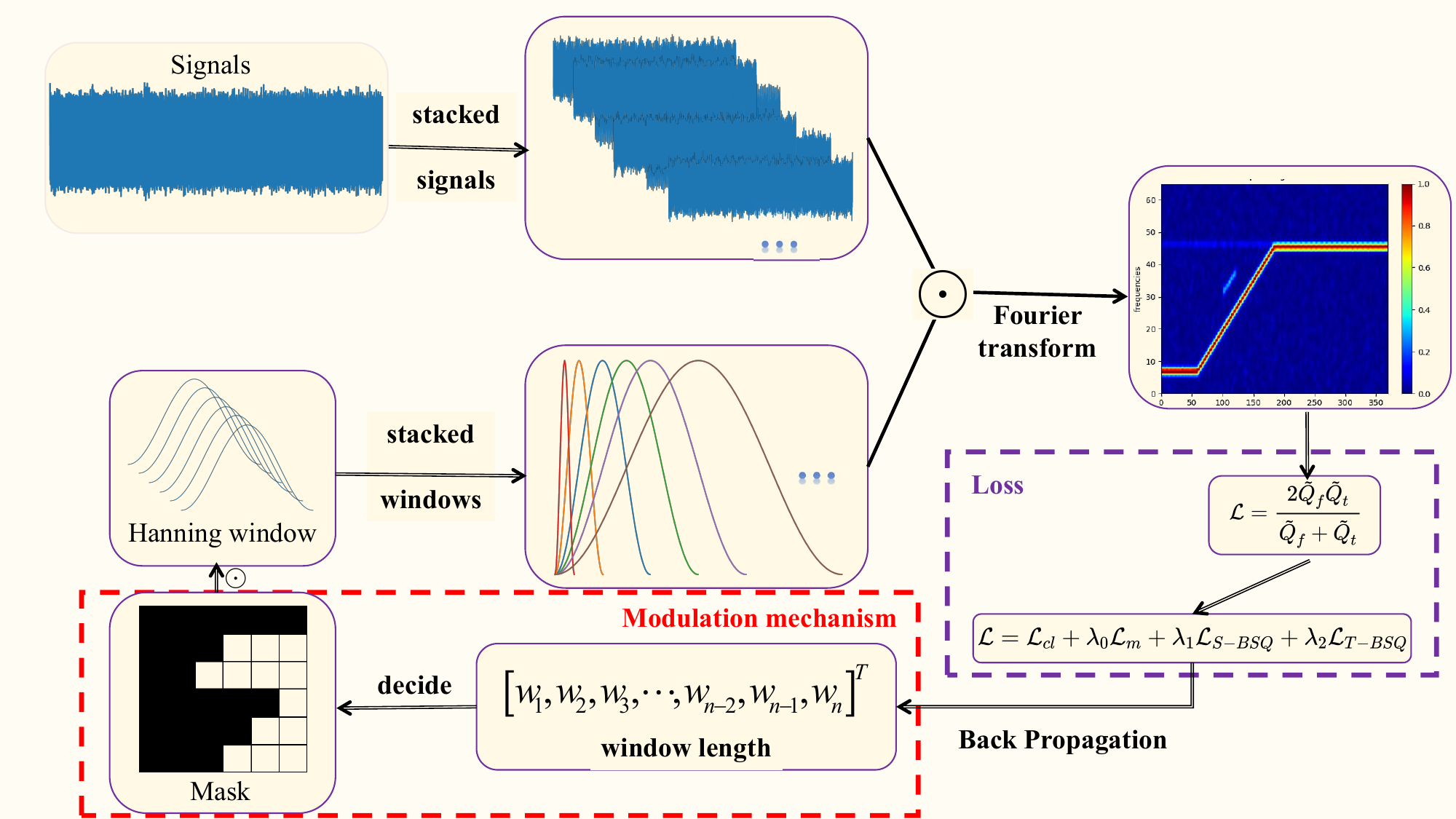}
  \caption{$\odot$ is element-wise product. Multiplying the signals with the window functions with variable window lengths, followed by DFFT yields MDSTFT. Subsequently,  the balanced spectrum quality loss is calculated, the optimal window function  is determined employing three evaluation metrics and backpropagation procedure.}
  \label{04}
\end{figure*}
\subsection{Physics-informed Balanced Spectrum Quality}
Intelligent spectrogram is emerging as an analytical tool for assessing the quality of the time-frequency spectrum. Nonetheless, as a measurement of the physical property of time-frequency resolution, this approach suffers from three significant drawbacks: \textbf{(\romannumeral1):} Intelligent spectrogram is only an evaluation indicator and cannot serve as a measure for the differentiable loss of the produced spectrogram within neural networks. \textbf{(\romannumeral2):} There exist dual methodologies for quantifying the intelligence spectrogram, such as frequency-domain quality coefficient ($\tilde Q_f$) and time-domain quality coefficient ($\tilde Q_t$). For some generated spectrogram, it is inconsistent rankings in $\tilde Q_f$ and $\tilde Q_t$, causing confusion for Engineers to select an appropriate parameter. As an outcome, there is remains a scarcity of a holistic evaluation index capable of amalgamating both methodologies. \textbf{(\romannumeral3):} The larger $\tilde Q_f$ and $\tilde Q_t$, the higher quality of the generated spectrogram, which contradicts the tenets of propelling the backpropagation algorithm towards minimal values.

Adhering to these issues leads us to propose the innovative solution--balanced spectrum quality loss, which helps to balance the time-frequency resolution of TFS and enhance energy aggregation, aiming at enhancing physical expression.

The physics-informed loss BSQ, serving as a regularization term, assists in mitigating over-fitting and diminishes the reliance on vast training datasets. This enhances the robustness of the trained model. It harnesses unlabeled data, facilitating the extraction of more generalized domain-invariant features. Furthermore, it ensures that the output of neural networks aligns more closely with physical principles, thus amplifying the interpretability of the output. BSQ not only steers MDSTFT in capturing high-quality TFS but also functions as a physical constraint, guiding learning domain-invariant features to augment the accuracy of cross-machine transfer diagnosis.

The quality coefficients serve as the proxies for two physical properties----temporal resolution and frequency resolution. As an initial step, compute the mean value of all rows and columns within the spectrogram:
\begin{flalign}\label{eq15}
&{\mu _f}\left[ t \right] = \frac{1}{F}\sum\limits_{f = 0}^{F - 1} {\left| {{\hat F}_\varpi }\left[ {t,f} \right] \right|} &
\end{flalign}
\begin{flalign}\label{eq16}
&{\mu _t}\left[ f \right] = \frac{1}{T}\sum\limits_{t = 0}^{T - 1} {\left| {{\hat F}_\varpi }\left[ {t,f} \right] \right|} &
\end{flalign}
where ${\mu _t}\left[ f \right]$ and ${\mu _f}\left[ t \right]$ signify the average values of time and frequency variables, respectively.

Following, calculate their standard deviation (${\sigma _f}\left[ t \right]$) and (${\sigma _t}\left[ f \right]$), as shown in \myref{eq17} and \myref{eq18}.
\begin{flalign}\label{eq17}
&{\sigma _f}\left[ t \right] = \sqrt {\frac{1}{{F - 1}}\sum\limits_{f = 0}^{F - 1} {{{\left| {\left| {{\hat F}_\varpi }\left[ {t,f} \right] \right| - {\mu _f}\left[ t \right]} \right|}^2}} } &
\end{flalign}
\begin{flalign}\label{eq18}
&{\sigma _t}\left[ f \right] = \sqrt {\frac{1}{{T - 1}}\sum\limits_{t = 0}^{T - 1} {{{\left| {\left| {{\hat F}_\varpi }\left[ {t,f} \right] \right| - {\mu _t}\left[ f \right]} \right|}^2}} } &
\end{flalign}

The balanced coefficient of variation is designated as the quotient of the mean divided by the standard deviation, which is not necessarily better as it is larger, but better as it is smaller. This corresponds precisely to the direction of minimization pursued in the backpropagation algorithm and constitutes an appropriate loss function during the optimization process.
\begin{flalign}\label{eq19}
&{{\tilde c}_f}\left[ t \right] = \frac{{{\mu _f}\left[ t \right]}}{{{\sigma _f}\left[ t \right]}}&
\end{flalign}
\begin{flalign}\label{eq20}
&{{\tilde c}_t}\left[ f \right] = \frac{{{\mu _t}\left[ f \right]}}{{{\sigma _t}\left[ f \right]}}&
\end{flalign}

Finally, obtain the Rational Quality Coefficient of the given time-frequency spectrum:
\begin{flalign}\label{eq21}
&{\tilde Q_f} = \frac{1}{T}\sum\limits_{t = 0}^{T - 1} {\frac{{{{\tilde c}_f}\left[ t \right]}}{{{{\tilde c}_f}{{\left[ t \right]}_{\max }}}}}  \in (0,1) &
\end{flalign}
\begin{flalign}\label{eq22}
&{\tilde Q_t} = \frac{1}{F}\sum\limits_{f = 0}^{F - 1} {\frac{{{{\tilde c}_t}\left[ f \right]}}{{{{\tilde c}_t}{{\left[ f \right]}_{\max }}}}}  \in (0,1)&
\end{flalign}

As illustrated in \myref{eq21} and \myref{eq22}, to circumvent the occurrence of the phenomenon of gradient explosion or disappearance, with adjustment of $\tilde Q_f$ and $\tilde Q_t$ to a maximum value 1.0. In intelligent spectrogram, it is observed that for $\tilde Q_f$ and $\tilde Q_t$ of the specific spectrogram, there exists an inconsistent ranking for a certain window length. For example, for $n=8 $,  the ranking of $\tilde Q_f$ is first, but the ranking of $\tilde Q_t$ is at the end \cite{h39}. To remedy this predicament, a novel loss function based on harmonic mean is devised, which incorporates both time resolution and frequency resolution into consideration.

In adherence to the principles of harmonic series, $\mathbcal{L}_{BSQ}$ is dictated by $\tilde Q_f$ and $\tilde Q_t$. The value of the harmonic mean (${\mathbcal{L}_{BSQ}}$) resides within the confines between $\tilde Q_f$ and $\tilde Q_t$. If ${\mathbcal{L}_{BSQ}}$ diminish, it is imperative for both $\tilde Q_f$ and $\tilde Q_t$ to be contemporaneously decreased. The minimization of $\mathbcal{L}_{BSQ}$ implies that $\tilde Q_f$ and $\tilde Q_t$ are also optimized towards the minimal direction. Hence, the interplay between the two was holistically taken into account for achieving the optimal window function optimization result. According to the definition of harmonic mean, \myref{eq23} can emerge:
\begin{flalign}\label{eq23}
&\begin{array}{l}
{\mathbcal{L}_{BSQ}} = {\left( {\frac{{{{\tilde Q}_f}^{{\rm{ - }}1}{\rm{ + }}{{\tilde Q}_t}^{{\rm{ - }}1}}}{2}} \right)^{{\rm{ - }}1}}{\rm{ = }}{\left( {\frac{{\frac{{{{\tilde Q}_f}{\rm{ + }}{{\tilde Q}_t}}}{{{{\tilde Q}_f}{{\tilde Q}_t}}}}}{2}} \right)^{ - 1}}\\
{\rm{ = }}\frac{{2{{\tilde Q}_f}{{\tilde Q}_t}}}{{{{\tilde Q}_f} + {{\tilde Q}_t}}}
\end{array}&
\end{flalign}

 BSQ analysis can balance two physical properties: temporal resolution and frequency resolution.
\subsection{Training pipeline of pyDSN for cross-machine transfer diagnosis}\label{section:033}
\begin{figure*}[h]
  \center
  \includegraphics[scale=0.5]{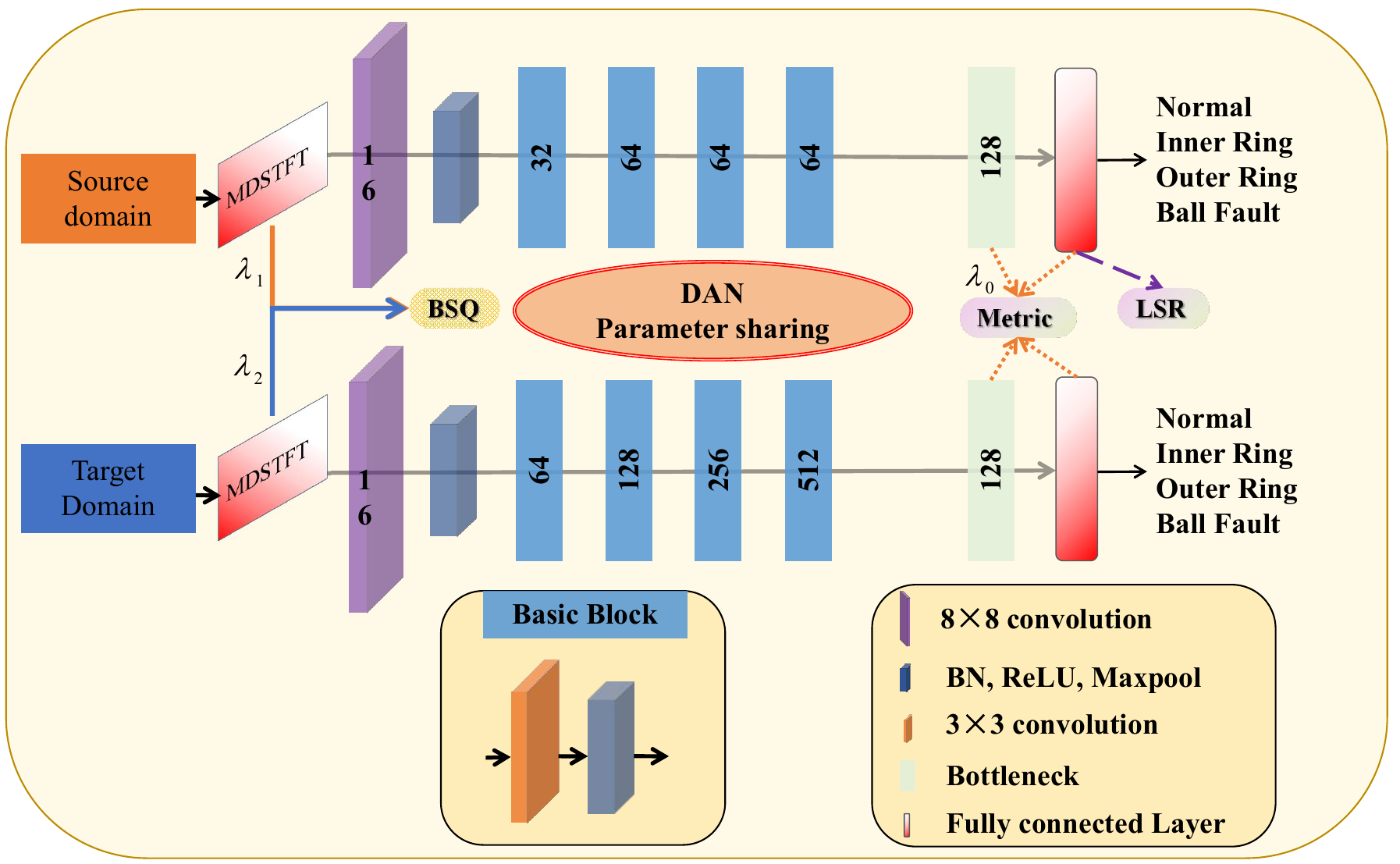}
  \caption{Flowchart of a PyDSN model.}
  \label{05}
\end{figure*}
The diagnostic procedure exhibited by pyDSN is illustrated in \cref{05}. During the forward propagation phase, variable speed data are fed into MDSTFT and DAN to generate soft labels, and subsequently, AdamW algorithm is deployed to optimize the parameters within pyDSN.
\begin{flalign}\label{eq25}
&\left\{ {\begin{array}{*{20}{c}}
{\begin{array}{*{20}{c}}
{\mathbcal{L} = {\mathbcal{L}_{cl}} + {\lambda _0}{\mathbcal{L}_m} + \lambda_1 {\mathbcal{L}_{S - BSQ}} + {\lambda _2}{\mathbcal{L}_{T - BSQ}}}\\
{{\Theta _s} \leftarrow {\Theta _s} - \mu (\frac{{\partial \mathbcal{L}}}{{\partial {\Theta _s}}})}
\end{array}}\\
{{\Theta _t} \leftarrow {\Theta _t} - \mu (\frac{{\partial \mathbcal{L}}}{{\partial {\Theta _t}}})}\\
{{\theta _0} \leftarrow {\theta _0} - \kappa  (\frac{{\partial \mathbcal{L}}}{{\partial {\theta _0}}})}\\
{{\theta _1} \leftarrow {\theta _1} - \kappa (\frac{{\partial \mathbcal{L}}}{{\partial {\theta _1}}})}
\end{array}} \right.&
\end{flalign}
where $\mu=0.001, \kappa=100.0$ are the learning rate, and $\Theta _s,\Theta _t,\theta _0,\theta _1$ are the hyper-parameters of the source domain classifier, target domain classifier, source domain MDSTFT, and target domain MDSTFT, respectively.

The proposed BSQ loss serves not only to guide the adjustment of STFT window lengths, but also to bind DAN. The reverse gradient propagation of BSQ can be representationally encapsulated as \myref{eq26}:
\begin{flalign}\label{eq26}
&\begin{array}{l}
\frac{{\partial {L_{BSQ}}}}{{\partial \Theta }} = \frac{{\partial ({{\left( {\frac{{{{\tilde Q}^{ - 1}}_f + {{\tilde Q}^{ - 1}}_t}}{2}} \right)}^{ - 1}})}}{{\partial \Theta }}\\
 =  - {\left( {\frac{{{{\tilde Q}^{ - 1}}_f + {{\tilde Q}^{ - 1}}_t}}{2}} \right)^{ - 2}}\left[ {\frac{{\partial {{(2{{\tilde Q}_f})}^{ - 1}}}}{{\partial \Theta }} + \frac{{\partial {{(2{{\tilde Q}_t})}^{ - 1}}}}{{\partial \Theta }}} \right]
\end{array}&
\end{flalign}
where $\Theta$ is the hyper-parameters of pyDSN.

The flowchart depicting the pseudo code for the total training of this model is shown as \myyyyyref{alg:01}.
\begin{algorithm*}[h]
\caption{The training and testing process of pyDSN.}
\label{alg:01}
\begin{algorithmic}[1]
\renewcommand{\algorithmicrequire}{\textbf{Input:}} 
\renewcommand{\algorithmicensure}{\textbf{Output:}} 
\Require
${D_s}={X^s_m}$: source domain; ${D_t}={X^t_m}$: target domain; ${\lambda_0}$,${\lambda_1}$,${\lambda_2}$: coefficients; $M$: training epoch;
\Ensure
Testing accuracy of PyDSN.
\State Initialize PyDSN
\For {$i \leftarrow 0;i \le M;i \leftarrow i + 1$}
\State Get source-domain spectrogram $S({D_s})$ from by MDSTFT.
\State Calculate source-domain balanced spectrum quality loss ${{\mathbcal{L}_{S - BSQ}}}$ by \myref{eq23}.
\State Get target-domain spectrogram $S({D_t})$ from BJTU-QS dataset by MDSTFT.
\State Calculate target-domain balanced spectrum quality loss ${{\mathbcal{L}_{T - BSQ}}}$ by \myref{eq23}.
\State Get source-domain feature $F({D_s})$.
\State Get source-domain soft prediction $P({D_s})$.
\State Calculate smoothed cross-entropy loss $\mathbcal{L}_{cl}$.
\State Get target-domain feature $F({D_t})$.
\State Get target-domain soft prediction $P({D_t})$.
\State Extract domain-invariant features using $\Im$.
\State Calculate domain metric loss $\mathbcal{L}_{m}$.
\State Train the parameters $\Theta_s$ of source-domain MDSTFT by ${{\mathbcal{L}_{cl}}}$, $\lambda_0{{\mathbcal{L}_m}}$, $\lambda_1{{\mathbcal{L}_{S - BSQ}}}$ and $\lambda_2{{\mathbcal{L}_{T - BSQ}}}$.
\State Train the parameters $\Theta_t$ of target-domain MDSTFT by ${{\mathbcal{L}_{cl}}}$, $\lambda_0{{\mathbcal{L}_m}}$, $\lambda_1{{\mathbcal{L}_{S - BSQ}}}$ and $\lambda_2{{\mathbcal{L}_{T - BSQ}}}$.
\State Train the feature extractor parameters ${\theta _1}$ by ${{\mathbcal{L}}}$.
\State Train the classifier parameters ${\theta _2}$ by ${{\mathbcal{L}}}$.
\EndFor
\State Save the trained model $\Theta $.
\State Get test data and load trained model $\Theta $.
\State Get test feature $F({D_t^t})$ and prediction $P({D_t^t})$.
\end{algorithmic}
\end{algorithm*}
\section{Application to wheelset Bearings of freight trains}\label{section:04}
\subsection{The description of datasets and tasks}
\subsubsection{Data description}
This investigation covers bearing and gearbox datasets: a dataset from Ottawa University; three datasets from the independently maintained experimental platform for bearings and gearboxes; and an additional dataset gathered from the real-world heavy haul freight train wheelset bearing platform designed and established by CRRC Qingdao Sifang. A total of five datasets are utilized to investigate the cross-mechine transfer diagnosis capabilities of the proposed pyDSN algorithm. A concise overview of these datasets is delineated below.
\begin{figure*}[h]
  \center
  \includegraphics[scale=0.4]{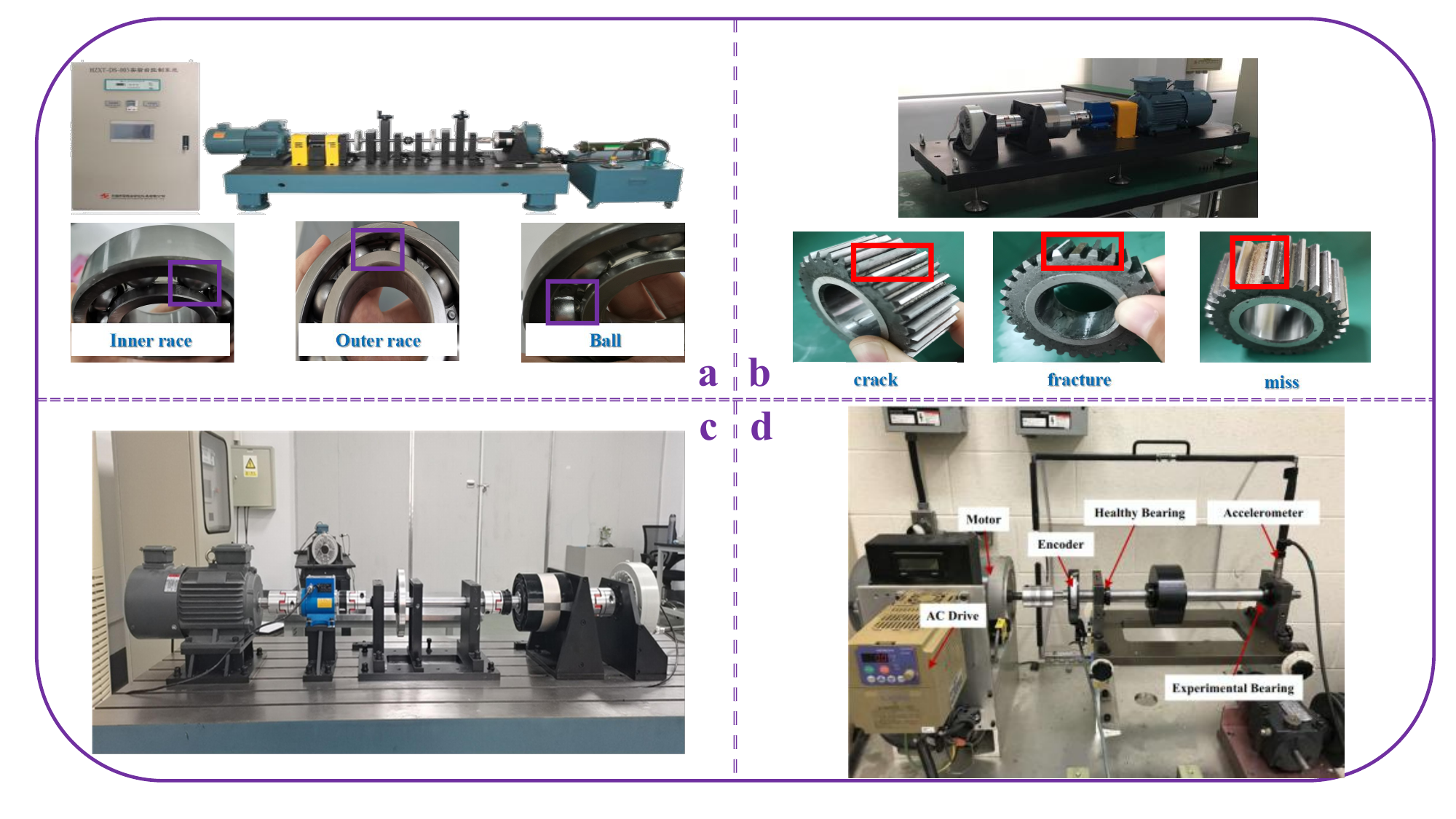}
  \caption{Equipments: (a) is double-span and double-rotor comprehensive fault experimentation platform; (b) is gearbox fault experiment platform; (c) is rotor-gear comprehensive fault experiment platform; (d) is experimental setup from University of Ottawa.}
  \label{06}
\end{figure*}
\begin{enumerate}[(1)]
\item $\rm{BJTU}_1$: The bearing fault dataset originates from the double-span and double-rotor comprehensive fault experimentation platform of Houde Automation Measurement as illustated in \cref{06}a. Utilized within the experiment is the NSK-6308 deep groove ball bearing at a sampling rate of 20 kHz. The experimental bearings into four categories based on their respective locations: Inner race fault (IR), Outer race fault (OR), Ball fault (B) and the healthy bearing (H). Localized pitting failures are mimicked through electrospark machining methods across these various sections.
\item $\rm{BJTU}_2$: The gearbox fault dataset comes from the planetary gearbox fault test bench of Houde Automation Measurement, as shown in \cref{06}b. The experimental setup, such as sensors and sampling frequency is identical with $\rm{BJTU}_1$. It simulates three distinct gearbox failures at varying rotational speeds, encompassing cracks, tooth fractures, tooth missing, and health.
\item $\rm{BJTU}_3$: The third bearing fault database originates from the rotor gear comprehensive fault experimental platform, depicted in \cref{06}c, where the employed rolling element model is the NSK-6205 single row deep groove ball bearing. The remaining experimental parameters adhere to those of $\rm{BJTU}_1$.
\item Ottawa \cite{h43}: The experimental setup of the variable speed rolling bearing dataset in the Mechanical Engineering Laboratory of the University of Ottawa, Canada is shown in \cref{06}d. The installation encompasses the ER16K ball bearing support shaft, replacing the selected experimental bearing with bearings presenting distinct health conditions. Furthermore, an accelerometer is positioned onto the external shell of the experimental bearing to acquire vibration data, with a sampling frequency 200 kHz.
\end{enumerate}

\begin{figure*}[h]
  \center
  \includegraphics[scale=0.4]{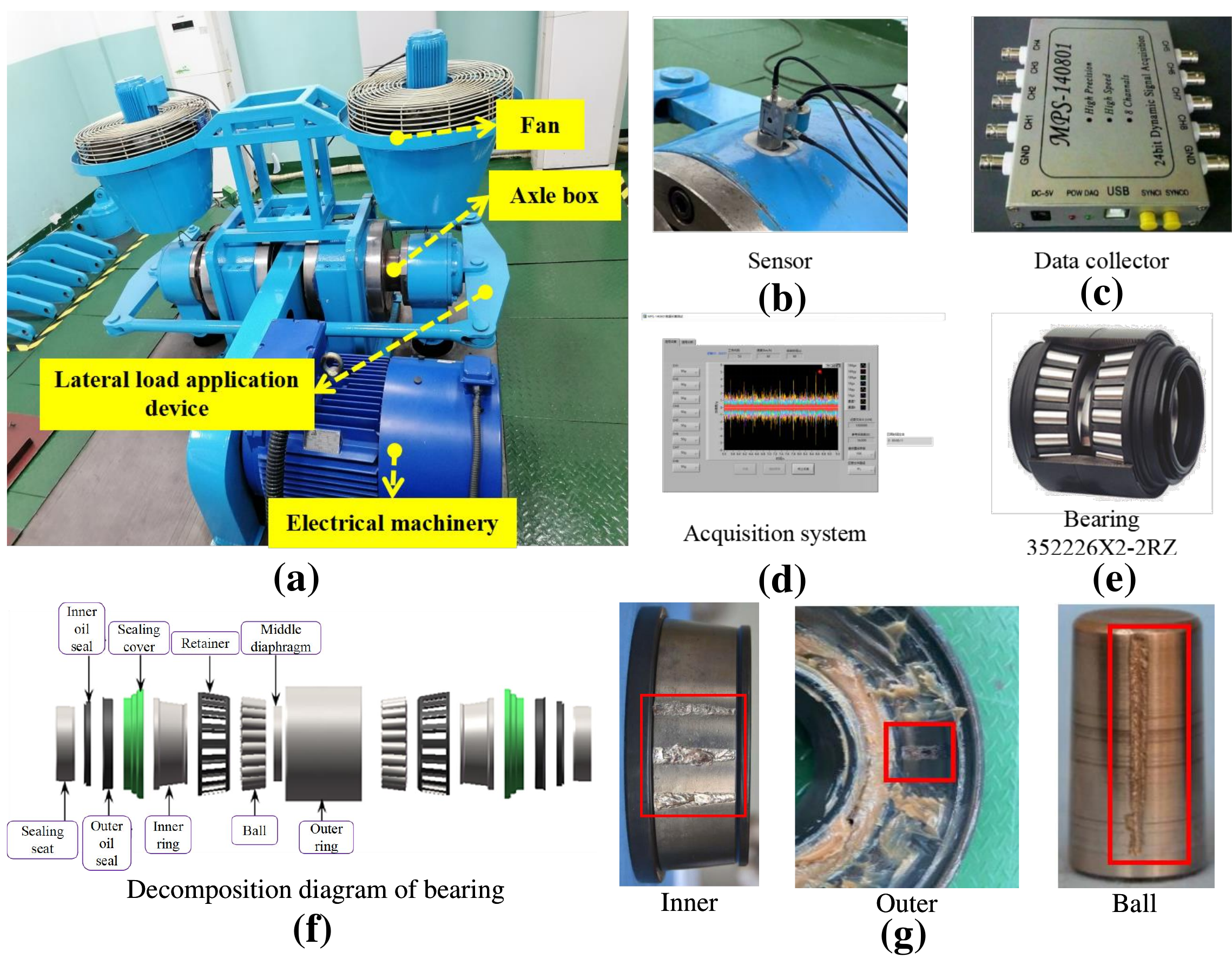}
  \caption{Heavy haul freight train wheelset bearing related equipment: (a) is heavy haul freight train wheelset bearing experimental platform; (b)is the data collector; (c) is the sensor; (d) is the acquisition system; (e) is the bearing (352226X2-2RZ); (f) is the schematic diagram of bearing disassembly; (g) is the partial failure types of railway freight train bearing.}
  \label{07}
\end{figure*}

The railway scenario dataset comes from the heavy haul freight train wheelset bearing platform developed by CRRC Qingdao Sifang, depicted in \cref{07}(a), which exhibits precision in simulating the payloads, axle bearing speeds, as well as the corresponding environment while under genuine operations, conforming to the real-world railway scenario. Specifically, vertical load is utilized to simulate axle load; Lateral load is used to simulate different line conditions, such as turning, passing through turnouts, and the lateral force on the track during uphill and downhill operations; The fan is employed to simulate crosswind during operation, setting between 8 m/s to 10 m/s. The standard bearings is equipped on one side of the wheelset and experimental bearing is mounted onto the opposite side. A vibration acceleration sensor is fixed on the outside of the axle box with a sampling frequency 16kHz. The tested bearing is 352226X2-2RZ (\cref{07}(e)(f)), which is a sealed dual row tapered roller bearing with double inner rings and middle isolation. The two end surfaces of the bearing are equipped with skeleton rubber sealing rings, with an inner diameter of 130 mm. It is used for accelerating freight car bogies and consists of outer rings, inner rings, rollers, cages, middle separators, sealing devices and other components. Three types of faults consist of: inner ring, outer ring, and balling, which are confirmed after dismantling and inspecting in the online monitoring system or manual check, In addition, it includes healthy bearings (H). The bearing rotation speed is set at $ 60km/h\sim 120 km/h $.

\begin{table}[htbp]
  \centering
  \caption{The detailed description of datasets.}
    \begin{tabular}{ccccc}
    \toprule
    Name  & Data Source & Speed(r/min) & Sampling frequency & Health status \\
    \midrule
    A     & BJTU\_1 & 900$ \sim $2400 & \multirow{2}[2]{*}{20kHz} & \multirow{2}[2]{*}{H,IF,OF,BF} \\
    A     & BJTU\_1 & 2400$ \sim $900 &       &  \\
\cmidrule{5-5}    B     & BJTU\_2\_bearing & 1000$ \sim $2500 & \multirow{4}[4]{*}{20kHz} & \multirow{2}[2]{*}{H,IF,OF,BF} \\
    B     & BJTU\_2\_bearing & 2500$ \sim $1000 &       &  \\
\cmidrule{5-5}    C     & BJTU\_2\_gear & 1000$ \sim $2500 &       & \multirow{2}[2]{*}{H,C,TF,TM} \\
    C     & BJTU\_2\_gear & 2500$ \sim $1000 &       &  \\
\cmidrule{5-5}    D     & Ottawa & accelerate & \multirow{2}[2]{*}{200kHz} & \multirow{2}[2]{*}{H,IF,OF,BF} \\
    D     & Ottawa & decelerate &       &  \\
\cmidrule{5-5}    E     & BJTU\_3 & 500$ \sim $2500 & \multirow{2}[2]{*}{20kHz} & \multirow{2}[2]{*}{H,C,TF,TM} \\
    E     & BJTU\_3 & 2500$ \sim $500 &       &  \\
\cmidrule{5-5}    F     & Heavy wagon & 60$ \sim $120 & \multirow{2}[2]{*}{16kHz} & \multirow{2}[2]{*}{H,IF,OF,BF} \\
    F     & Heavy wagon & 120$ \sim $60 &       &  \\
    \bottomrule
    \end{tabular}%
  \label{tab01}%
\end{table}%

\subsubsection{Task description}
The heavy haul freight train inevitably encounters acceleration and deceleration during operation, which may harbor distinctive features compared to constant speed data. By utilizing self-collected and public variable speed data, the heavy haul freight train wheel bearing health monitoring is addressed through cross-machine transfer diagnosis methodology.

The signal is segmented into samples using a sliding window, with a sample length 3072. In the scenario on time-varying speeds, there are 100 samples for each category in cross-machine transfer diagnosis. Therefore, each category encompasses 100K samples in source domain or target domain. The test set comes from the target domain, yielding a total collection of 900K samples (K represents the number of categories). The multiple cross-machine tasks are constructed, where A$\rightarrow$F represents A as the source domain and F as the target domain. All relevant experiments are executed on RTX 4090 24GB GPU equipped with Pytorch 2.0.

The experiment are repeated five times and record the average and standard deviation. The other parameter settings are shown in \myyyref{tab02}. In addition to the bearing dataset, in order to improve the credibility and generalization, gearbox data is employed to validate the performance with pyDSN. Of these, A, B, D, and F are bearing datasets. C and E are gearbox datasets.
\begin{table}[htbp]
  \centering
  \caption{Model parameter settings}
    \begin{tabular}{cc}
    \toprule
    Parameter & Value \\
    \midrule
    batch\_size & 20 \\
    Backbone & WDCNN \\
    window function & Kaiser \\
    Optimizer & AdamW \\
    stride & 16 \\
    max\_epoch & 200 \\
    seed  & 3407 \\
    training set & 400 \\
    testing set & 3600 \\
    lr    & 0.001 \\
    lr (MDSTFT) & 100.0 \\
    \bottomrule
    \end{tabular}%
  \label{tab02}%
\end{table}%
\subsection{Case analysis}
\subsubsection{Comparison performance with DSN, STFT, and pyDSN}
\begin{figure}[h]
  \center
  \includegraphics[scale=0.27]{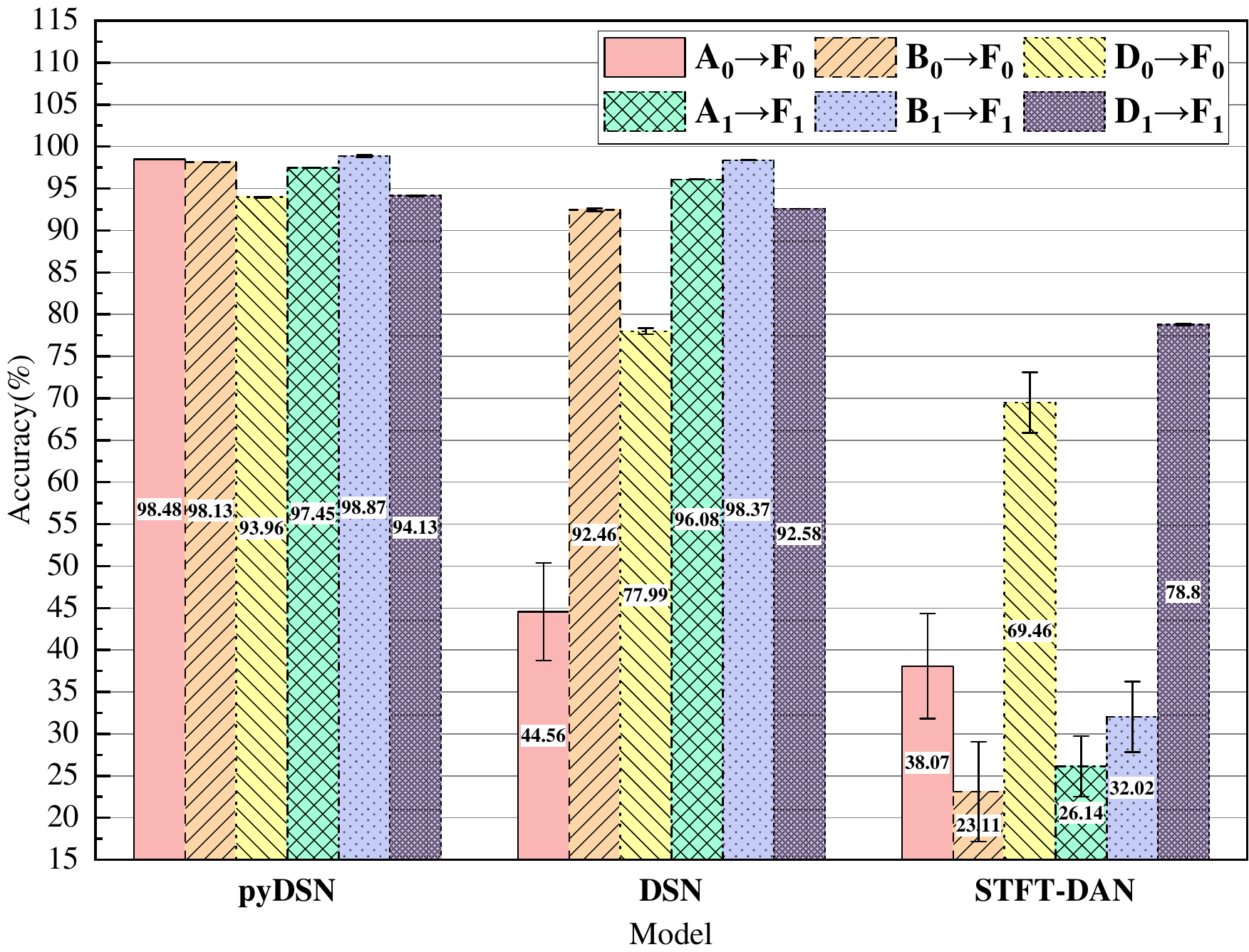}
  \caption{Performance with three methods of different transfer tasks on bearings.}
  \label{08}
\end{figure}

pyDSN is improved over the typical STFT and DSN. To evaluate the health status of wheelset bearings, the wheelset bearing dataset is applied as the target domain, while other datasets are utilized as the source domain. As shown in \cref{08}, among the six acceleration and deceleration tasks executed, pyDSN achieved an average accuracy of 96.85\%, far surpassing DSN (83.67\%). This is due to the fact that MDSTFT does not utilize a fixed window length, but instead uses multiple different windows, which can adaptively collect fault information on conditions at variable speeds. Moreover, pyDSN, under the influence of BSQ loss, enables MDSTFT to generate high-quality spectrograms, which helps to mitigate the distribution shifts between the source and target domains. The two-stage strategy, STFT-DAN, which uses labor experience to select window sizes, has limited performance. DSN with a fixed differentiable window faces a larger distribution shifts between domains, which makes it off-limits to extract discriminative fault feature information, resulting in an accuracy of less than 80\%.

\begin{figure}[h]
  \center
  \includegraphics[scale=0.25]{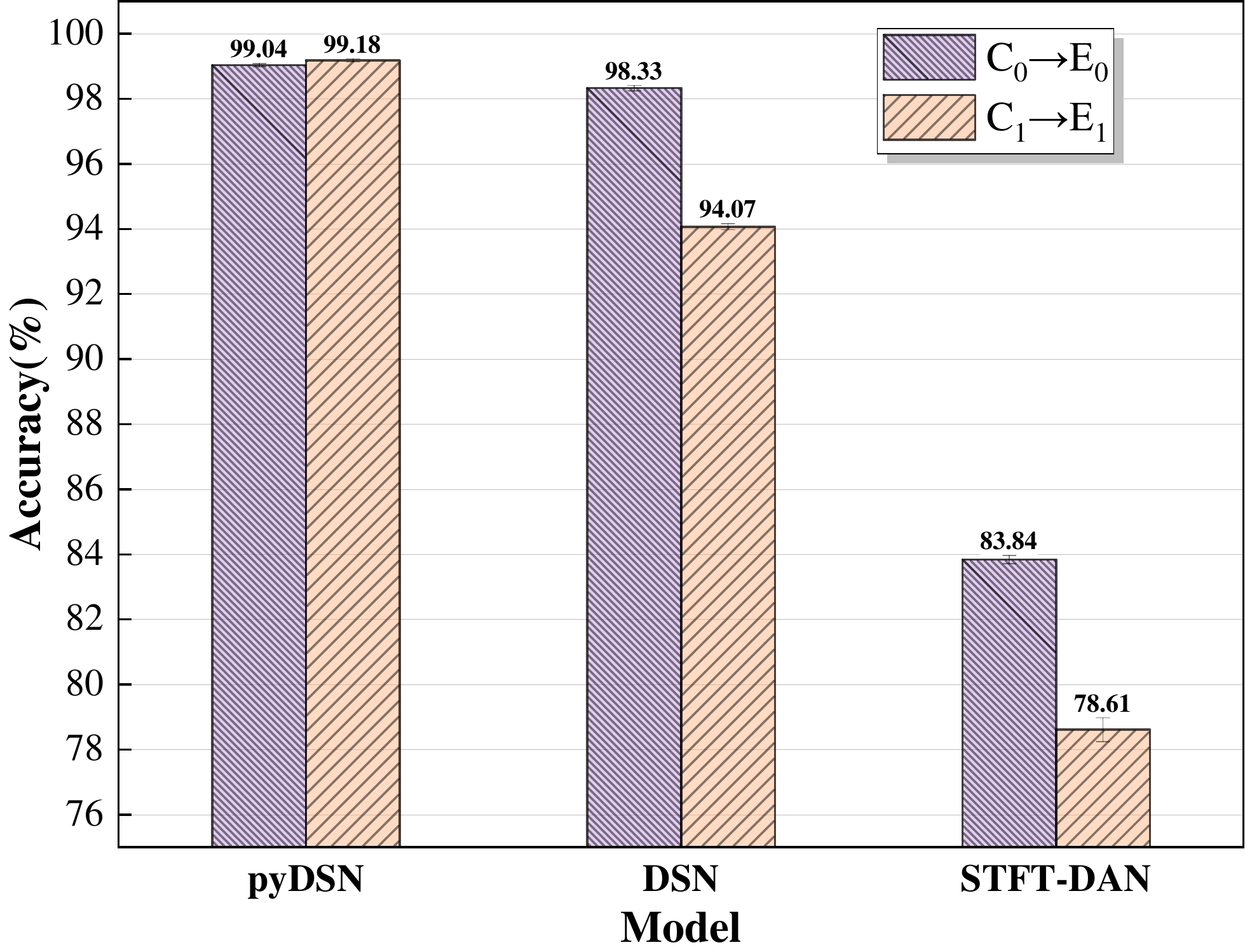}
  \caption{Performance with three methods of different transfer tasks on gearboxes.}
  \label{09}
\end{figure}

To verify comprehensive generalization capability of pyDSN, as shown in \cref{09}, experiments are conducted on the gearbox components. B and D are collected in-house. Compared pyDSN and DSN, self-collected datasets generally exhibit a high signal-to-noise ratio and obvious fault characteristics, making it relatively easy to obtain domain-invariant features. Despite pyDSN (99.11\%) and DSN (96.20\%) with high diagnostic accuracy ($>$96.00\%), pyDSN still improves by 2.91\% in comparison to DSN. STFT-DAN (81.23\%) still performs poorly, highlighting the significance of data enhancement in improving transferability under speed fluctuations and small sample conditions. In summary, although signal processing universally acknowledged as a preprocessing methodology, differentiable signal processing  substantially enhances the data feature enhancement capacity of signal processing algorithms due to its robust ability to adaptively adjust pivotal parameters with the assistance of gradient-descent algorithms.
\subsubsection{The plug-and-play MDSTFT and BSQ}
As illustrated in \myyyref{tab03}, taking $A_0\rightarrow F_0$ as an example, BSQ, MDSTFT, and DSTFT can all be seamlessly integrated into several classic networks without conflicts. We have selected lightweight model (EfficientNet v2), deep models (ResNet-18 and DenseNet-121), as well as fault diagnosis models (WDCNN and DRSN-CW), as these models are extensively utilized and representative in fault diagnosis. BSQ (w/o) signifies the utilization of MDSTFT, without BSQ loss; MDSTFT represents the usage of BSQ loss; DSTFT denotes the usage of DSTFT and BSQ loss.

Faced with demanding cross-machine and variable speed condition tasks, the performance with DSTFT depends on the integrated backbone, exhibiting significant difference in performance under different backbones. For example, ResNet-18 can achieve 83.95\%, while it drops to 47.82\% when utilizing lightweight DRSN-CW, indicating poor generalization when processing speed fluctuation datasets. MDSTFT can improve this by 40.76\% compared with DSTFT, suggesting that employing different windows can effectively extract exclusive fault information. By comparing whether to use physics-informed BSQ, the hybrid-driven pyDSN improves by 11.09\%, implying that the hybrid-driven model incorporating physical limitations can enhance the capability to obtain discriminative features. This proves that the physics-informed regularization processes a strong driving force for extracting domain-invariant features and improving the generalization capacity.
\begin{table}[b]
  \centering
  \caption{Different backbones being integrated BSQ and MDSTFT.}
    \begin{tabular}{cccc}
    \toprule
    Model & BSQ(w/o) & DSTFT & MDSTFT \\
    \midrule
    EfficientNet v2 \cite{h49} & 85.56\% & 43.46\% & 91.61\%($\uparrow6.05\%$) \\
    ResNet-18 \cite{h50} & 88.69\% & 83.95\% & 95.83\%($\uparrow7.14\%$) \\
    DenseNet-121 \cite{h51}  & 70.86\% & 59.55\% & 90.92\%($\uparrow20.06\%$) \\
    WDCNN-2D \cite{h52}  & 94.56\% & 33.21\% & \cellcolor{gray!50}97.31\%($\uparrow2.75\%$) \\
    DRSN-CW \cite{h53} & 76.67\% & 47.82\% & 96.11\%($\uparrow19.44\%$) \\
    \rowcolor{gray!20}
    Average & 83.27\% & 53.60\% & \textbf{94.36\%($\uparrow11.09\%$)} \\
    \bottomrule
    \end{tabular}%
  \label{tab03}%
\end{table}%
\begin{figure}[h]
  \center
  \includegraphics[scale=0.3]{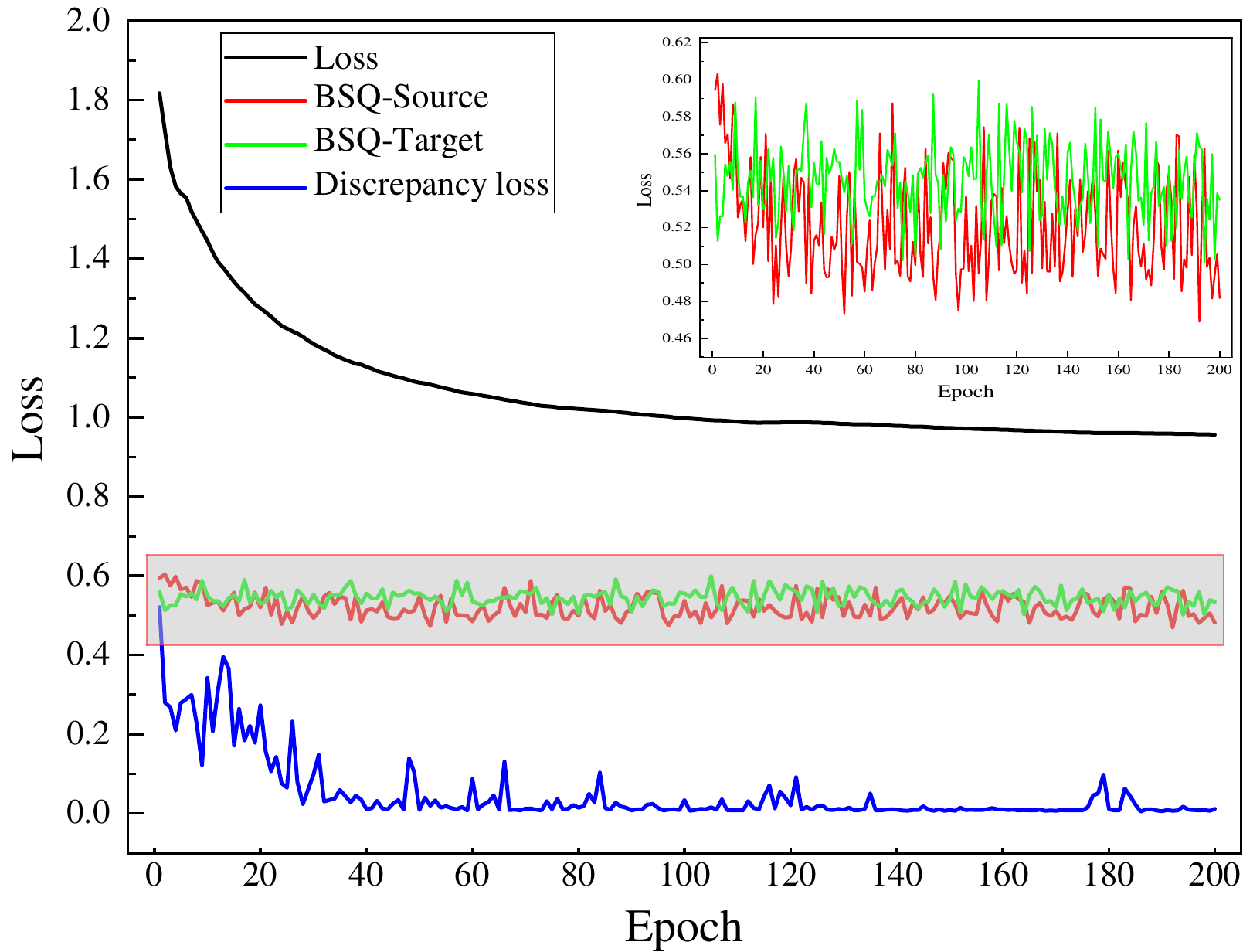}
  \caption{The diagram of loss function iteration.}
  \label{10}
\end{figure}

As shown in \cref{10}, the iterative process of the loss function is delineated during the learning phase. Loss represents the overall loss function (\myref{eq24}), and joint maximum mean discrepancy (JMMD) loss (${\mathbcal{L}_m}$) is identified to measure the discrepancy between source and target domains. At the conclusion of iteration, both losses ultimately attain full convergence. In addition, the BSQ loss in the source domain decreased from 0.59438 to 0.46898, whilst the BSQ loss associated with the target domain decreased from 0.55964 to 0.50112. This observation highlights that the gradient-descent algorithm can effectively search for the optimal window lengths, enhance spectral energy clustering, and further facilitate the learning of domain-invariant features.
\subsubsection{The experiment of BSQ}
To demonstrate the pioneering nature of the proposed BSQ. A comprehensive evaluation is conducted using prevalent physics-informed losses: Entropy and Kurtosis are the predominantly metrics for evaluating the spectrogram energy focusability; BSQ is a trailblazing iteration from Intelligent Spectrum; Cycle-Stationarity \cite{h40} is another method. $\tilde Q_f$ only focuses on frequency resolution, while $\tilde Q_t$ only considers time resolution, while BSQ aims to strike a balance between these two dimensions. The accuracy stands at 97.25\%, when jointly considering time-frequency resolution, respectively yielding 2.42\% and 4.36\% over either consideration alone. This underscores that the spectrogram needs to balance time-frequency resolution to accurately represent fault information.

Moreover, compared to other physics-guided losses, BSQ also showcases superior performance, significantly improving the adaptability to cross-domain, with a 9.64\% increase over Cycle-Stationarity. BSQ proves to a highly efficacious assessment tool to quantify the energy distribution and fault feature expression within spectrograms.
\begin{figure}[h]
  \center
  \includegraphics[scale=0.3]{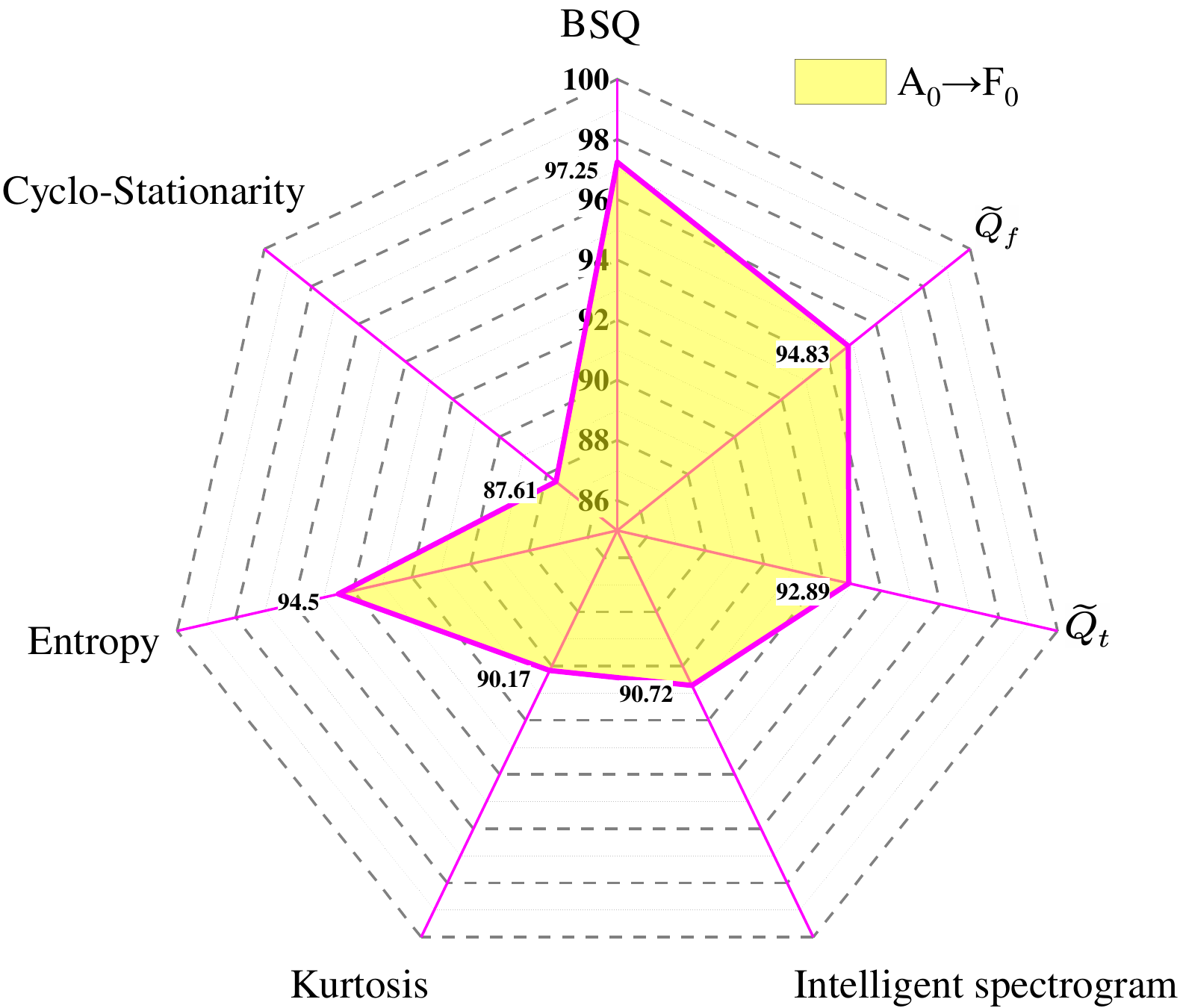}
  \caption{The accuracy comparison with other loss functions (\%).}
  \label{11}
\end{figure}
\subsection{The performance analysis of pyDSN on heavy haul freight train wheelset bearings}
To assess the superior performance of pyDSN with other state of the arts (SOTA). The first type is the classical mapping-based  models such as DDC (MMD), DCORAL, DJDA, and DSAN; The second type comprises the adversarial domain adaptation model DANN; The third type consists of emerging models the emerging models like CK-MMD \cite{h57}, MMSD \cite{h55}, and DDTLN \cite{h56} for cross-machine transfer diagnosis; The last type is the con-temporarily popular hybrid-driven deep models embedded signal knowledge, where signal processing is regarded as a form of data augmentation, and these models have achieved remarkable performance in their respective tasks, encompassing SincNet \cite{h58}, MCN\_WFK \cite{h831}, NTScatNet \cite{h59}, DFAWNet \cite{h60}, GTFENet \cite{h61}, GaborNet \cite{h62}, WIDAN\cite{h67}, MWPN \cite{h63}, and TFN \cite{h64}, as illustrated in \myyyref{tab04}.
\begin{table*}[htbp]
  \centering
  \caption{Performance comparison with other methods (\%).}
    \begin{tabular}{llllllll}
    \toprule
          & $A_0\rightarrow F_0$  & $B_0\rightarrow F_0$  & $D_0\rightarrow F_0$  & $A_1\rightarrow F_1$  & $B_1\rightarrow F_1$  & $D_1\rightarrow F_1$  & Average \\
    \midrule
    \rowcolor{blue!15}
    DDC   & 44.92±2.88 & 59.33±3.17 & 50.42±4.00 & 49.17±0.46 & 28.41±0.03 & 67.92±2.13 & 50.03 \\
    \rowcolor{blue!15}
    DCORAL & 32.17±2.75 & 33.33±2.41 & 37.58±1.04 & 29.17±0.05 & 28.33±0.09 & 63.08±1.25 & 32.28 \\
    \rowcolor{blue!15}
    DJDA  & 29.18±0.93 & 50.96±4.57 & 50.32±8.38 & 45.57±5.25 & 33.11±0.09 & 61.75±1.88 & 45.15 \\
    \rowcolor{blue!15}
    DSAN  & 32.07±1.34 & 33.21±0.86 & 41.33±0.83 & 47.75±6.17 & 27.33±0.13 & 67.42±0.59 & 41.52 \\
    \rowcolor{blue!15}
    DANN  & 30.47±0.36 & 36.19±0.60 & 61.00±0.61 & 43.19±0.59 & 18.69±0.03 & 47.61±0.31 & 39.53 \\
    \rowcolor{gray!20}
    CK-MMD \cite{h57} & 30.42±1.21 & 38.75±2.30 & 31.75±0.03 & 52.33±1.00 & 28.91±0.09 & 64.00±1.08 & 41.02 \\
    \rowcolor{gray!20}
    MMSD \cite{h55}  & 89.75±7.75 & 19.92±1.55 & 91.25±3.92 & 26.17±1.24 & 46.08±9.97 & 64.08±1.29 & 56.21 \\
    \rowcolor{gray!20}
    DDTLN \cite{h56} & 30.14±0.78 & 33.21±1.95 & 64.14±3.32 & 29.54±6.45 & 52.33±2.41 & 55.12±1.37 & 44.08 \\
    \rowcolor{blue!8}
    SincNet \cite{h58} & 98.67±0.34 & 33.08±1.24 & 67.25±1.46 & 65.08±5.45 & 27.17±0.51 & 76.00±0.77 & 61.21 \\
    \rowcolor{blue!8}
    MCN\_WFK \cite{h831} & 27.50±2.17 & 32.00±0.00 & 36.50±1.08 & 28.00±0.29 & 26.00±0.00 & 35.25±1.54 & 30.88 \\
    \rowcolor{blue!8}
    NTScatNet \cite{h59} & \textbf{100.00±0.00} & 71.00±2.54 & 72.61±0.72 & 84.54±1.36 & 46.67±2.38 & 71.38±0.50 & 74.37 \\
    \rowcolor{blue!8}
    DFAWNet \cite{h60} & 94.63±0.59 & 86.05±3.21 & 59.63±2.70 & 80.35±1.44 & 29.15±2.20 & 68.28±0.96 & 69.68 \\
    \rowcolor{blue!8}
    GTFENet \cite{h61} & 57.92±1.32 & 40.96±1.03 & 58.58±3.76 & 60.33±3.19 & 60.50±0.57 & 89.88±2.42 & 61.36 \\
    \rowcolor{blue!8}
    GaborNet \cite{h62}  & 39.75±0.79 & 57.50±1.55 & 30.17±4.19 & 32.25±1.14 & 49.00±2.95 & 32.75±1.22 & 40.23 \\
    \rowcolor{blue!8}
    WIDAN \cite{h67} & 96.88±5.81 & 22.13±4.21 & 70.63±4.86 & 71.88±1.30 & 72.92±1.85 & 59.38±0.87 & 65.64 \\
    \rowcolor{blue!8}
    MWPN \cite{h63}  & 50.17±1.48 & 28.08±0.84 & 75.58±0.71 & 58.58±2.13 & 44.83±0.60 & 65.18±4.14 & 53.74 \\
    \rowcolor{blue!8}
    TFN \cite{h64}  & 75.00±0.43 & 37.25±1.63 & 96.75±1.50 & 33.75±3.70 & 38.00±1.90 & 85.60±2.29 & 61.06 \\
    \rowcolor{gray!50}
    pyDSN & 94.81±3.02 & \textbf{96.64±0.42} & \textbf{98.78±0.50} & \textbf{93.83±0.29} & \textbf{98.86±0.32} & \textbf{99.00±0.04} & \textbf{96.97} \\
    \bottomrule
    \end{tabular}%
  \label{tab04}%
\end{table*}

The first three algorithms mentioned are purely data-driven, which perform inadequately under cross-machine variable speed tasks, which indicates that data-driven models depend on the quantity and quality of available data. In the presence of small samples and larger disparity between the source and target domains, these algorithms are likely to fail. Conversely, the hybrid-driven methods demonstrate highly competitive performance in the face of such challenging tasks. The signal processing module can reduce the noises, highlight fault information, and mitigate the adverse effects of noise on the source and target domains. However, reasonable signal processing methods are required to achieve this effect. For instance, utilization of Gabor transform results in an average accuracy of only 40.23\%, which is even worse than some purely data-driven methods.

The task encompasses two thorny challenges. One is across machines, as the data collected by different devices has greater distribution shifts; The other challenge is variable speed data, where the fault frequency is changing caused by changes in speed, rather than like constant speed. When pyDSN is tested against six tasks, the accuracy is 94.81\%, 96.64\%, 98.78\%, 93.83\%, 98.86\%, and 99.00\%, respectively, with an average accuracy is 96.97\%. This represents a minimum improvement of  22.7\%. The MDSTFT module in pyDSN aims to use differentiable windows and physical losses to extract fault information at variable speeds, while shortening the distribution differences between the source and target domains. However, other tasks perform poorly due to the lack of the aforementioned either.

The superior domain-invariant features can be explained from two perspectives. On one hand, domain-invariant features are extracted under the guidance of MDSTFT that furnishes an interpretable initialization input for acquiring domain-invariant features. Meanwhile, MDSTFT maintains interpretable with following the calculation process of STFT. It can be seamlessly integrated with domain adaptation networks without conflicts. On the other hand, domain-invariant features are extracted under the guidance of BSQ. BSQ measures energy focusability of TFS and balances two physical properties: time-frequency resolution. It directs the optimization trajectory of neural networks, allowing physical insights to influence the optimization process, and facilitates to extract more robust domain-invariant features, making the feature representations more closely with physical principles and enhancing interpretability.

According to Table \ref{tab04}, in the Task $A_0\rightarrow F_0$, interpretable methods such as SincNet and NTScatNet demonstrate superior performance. For the specific task, These methods utilize the Sinc function and wavelet scattering structure, respectively, to extract generalized features from the source domain, thereby minimizing the discrepancy between the two domains. In the case of pyDSN, this subpar performance could be attributed to the influence of regularization weight parameters, which may result in extracted features that do not effectively represent domain-invariant features. Nevertheless, this performance decline is within acceptable limits, with the accuracy remaining above 94.00\%. In contrast, in other tasks, SincNet and NTScatNet exhibit severe performance degradation, which underscores the stability, robustness, and best average precision of the proposed pyDSN.

\begin{figure*}
  \centering
	\begin{minipage}{0.33\linewidth}
		\centering
		\includegraphics[width=1.0\linewidth]{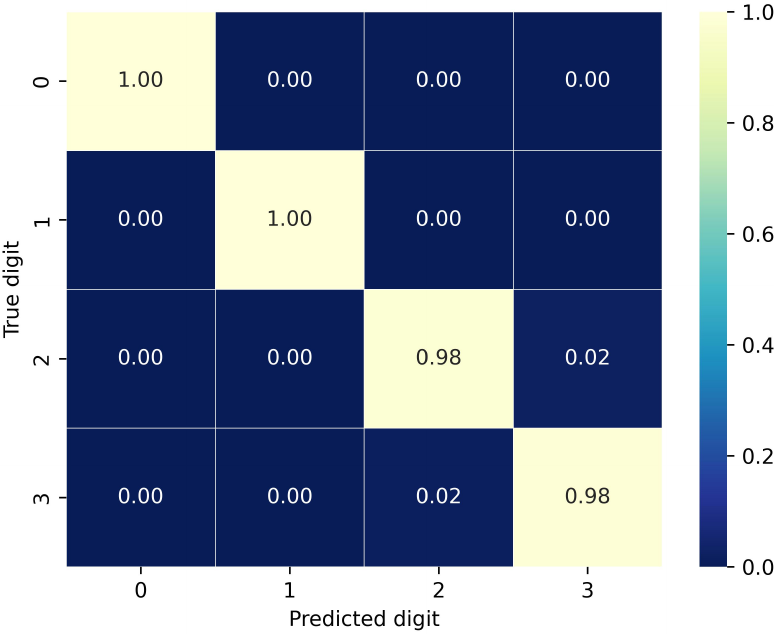}
		\subcaption{pyDSN.}
		\label{1200}
	\end{minipage}
	\begin{minipage}{0.33\linewidth}
		\centering
		\includegraphics[width=1.0\linewidth]{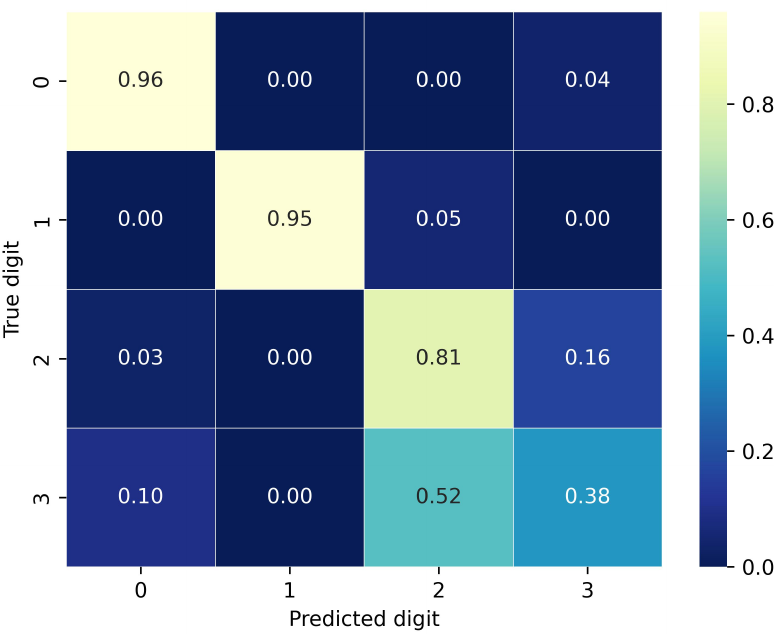}
		\subcaption{GTFENet.}
		\label{1201}
	\end{minipage}
    \begin{minipage}{0.33\linewidth}
		\centering
		\includegraphics[width=1.0\linewidth]{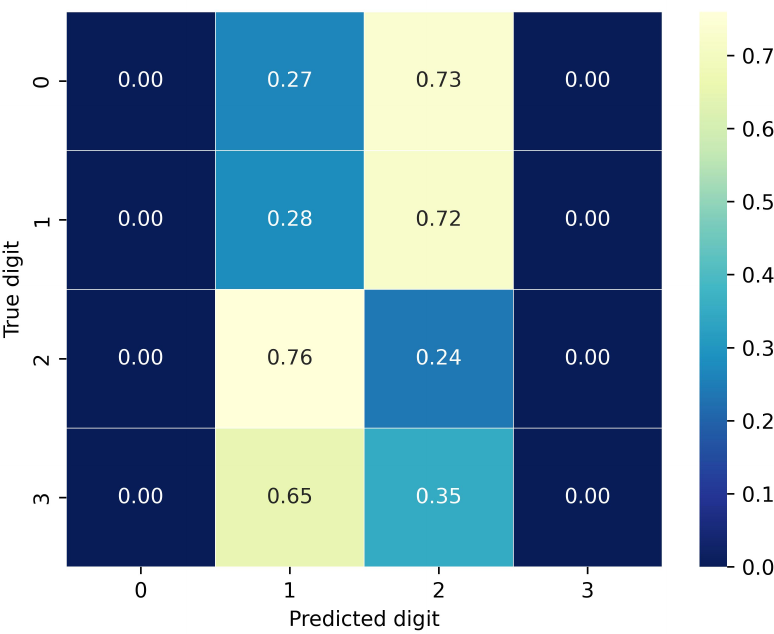}
		\subcaption{GaborNet.}
		\label{1202}
	\end{minipage}

    \begin{minipage}{0.33\linewidth}
		\centering
		\includegraphics[width=1.0\linewidth]{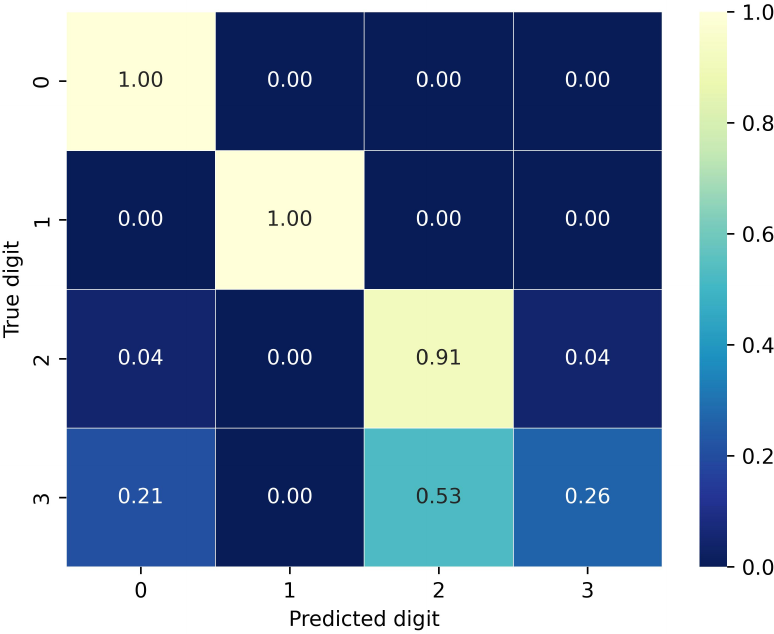}
		\subcaption{WIDAN.}
		\label{1203}
	\end{minipage}
    \begin{minipage}{0.33\linewidth}
		\centering
		\includegraphics[width=1.0\linewidth]{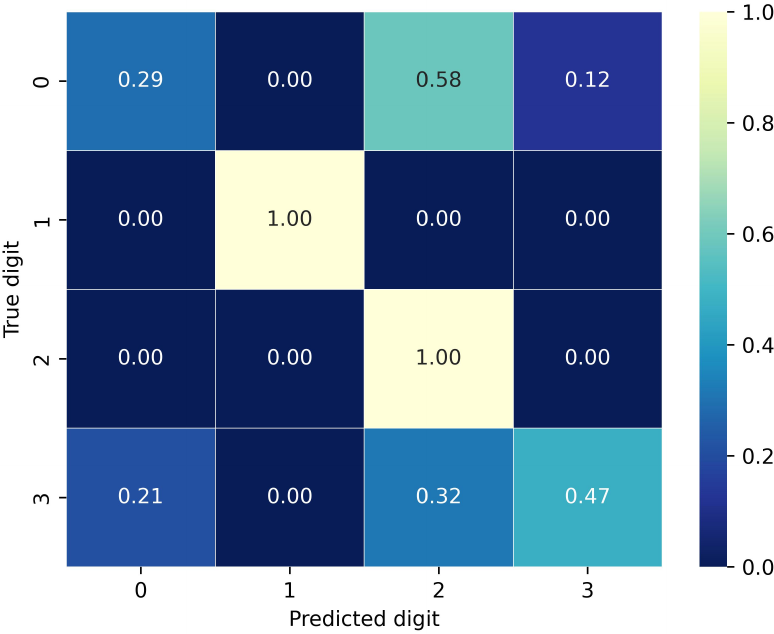}
		\subcaption{DFAWNet.}
		\label{1204}
	\end{minipage}
	\begin{minipage}{0.33\linewidth}
		\centering
		\includegraphics[width=1.0\linewidth]{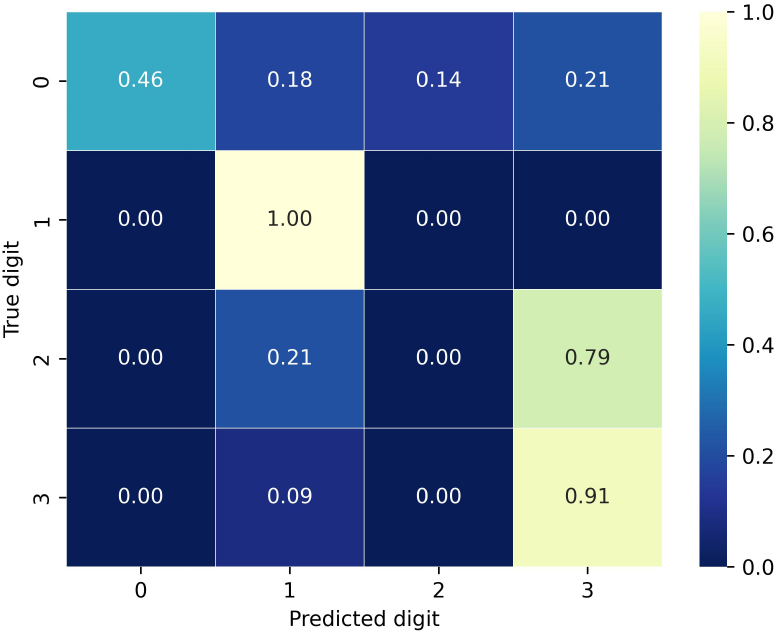}
		\subcaption{TFN.}
		\label{1205}
	\end{minipage}
  \caption{The confusion matrix of six methods on $D_1\rightarrow F_1$.}
  \label{12}
\end{figure*}

As shown in \cref{12}, the performance of pyDSN and other methods across $D_1\rightarrow F_1$ is evaluated through the confusion matrix. The confusion matrix can provide a comprehensive evaluation of different categories. Among these models, the most fundamental criterion is "Zero False Alarm" and "Zero Missed Alarm", and it is clear that only pyDSN can satisfy this requirement. In tasks, identifying inner ring and ball faults proves to be challenging, while other categories can recognize over 99\%. This finding serves as a valuable direction for future research endeavors.

\section{The interpretability analysis of MDSTFT}\label{section:05}
The Spectra Quest Variable (SQV) speed dataset, with a sampling frequency of 25.6kHz, is a publicly available dataset for variable speed analysis from Xi'an Jiaotong University. It is collected during a continuous speed change, encompassing a complete process of acceleration and deceleration. It begins from a stationary state, gradually increasing to 3000 rpm/min, where it stabilizes, and subsequently decelerates gradually back to zero. It serves as a good benchmark dataset and can be used as an example for research in this field. More details about the dataset can be found in \cite{h65}.

MDSTFT is considered inherently interpretable due to its unerring adherence to core principles of STFT and produces reliable results. Additionally, the learning rate of MDSTFT is 100.0, which is appreciably different from DAN. A larger learning rate serves to broaden the exploration spectrum of the time window and make it easier to find the optimal window lengths. In comparison to DSTFT, MDSTFT incorporates a signal modulation mechanism, which allows the raw signals to see a portion of the windows during FFT, thereby instigating different window lengths for capturing fault frequencies.

The rotational speed is continuously fluctuating. it is difficult to accurately measurement speed changes due to the high cost of tachometer. MDSTFT does not rely on the rotational speed to extract various fault features but achieves the same effect by using the time-varying window lengths under the guidance of the modulation mechanism. It does not require prior information such as rotational speed.

The calculation of bearing fault frequency necessitates prior knowledge of parameters including the rolling element diameter, ball pitch diameter, bearing contact angle, number of rolling elements, and rotational frequency. Not all devices can obtain these parameters, and the formulas are only applicable to constant rotational speed conditions. MDSTFT also do not need the information such as bearing dimensions.

Therefore, MDSTFT does not require prior knowledge of information such as speed and bearing size to obtain fault frequency.

The progressiveness of MDSTFT can be demonstrated from both qualitative and quantitative perspectives.
\subsection{Qualitative perspective}
\begin{figure*}
  \centering
	\begin{minipage}{0.33\linewidth}
		\centering
		\includegraphics[width=1.0\linewidth]{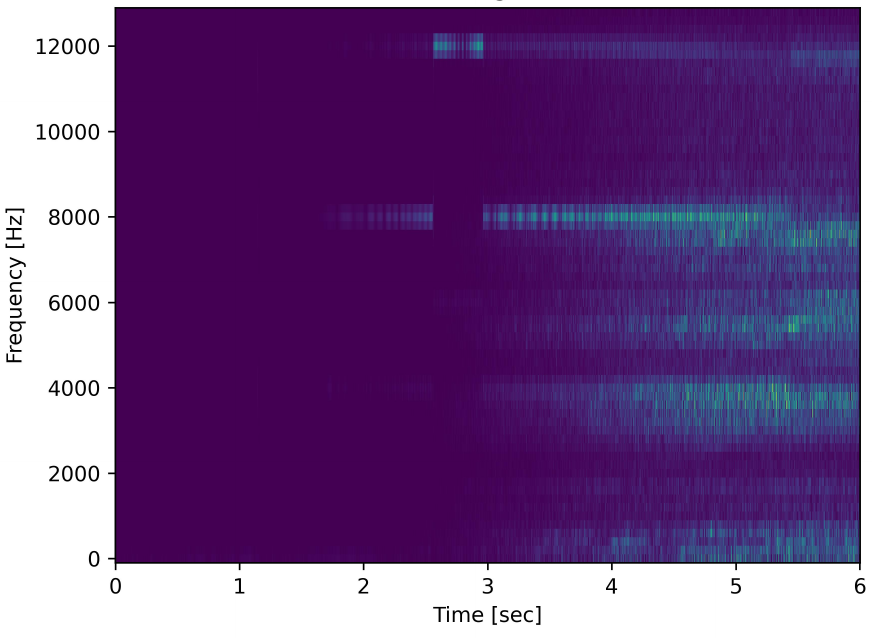}
		\subcaption{SQV\_acc\_STFT.}
		\label{1301}
	\end{minipage}
	\begin{minipage}{0.33\linewidth}
		\centering
		\includegraphics[width=1.0\linewidth]{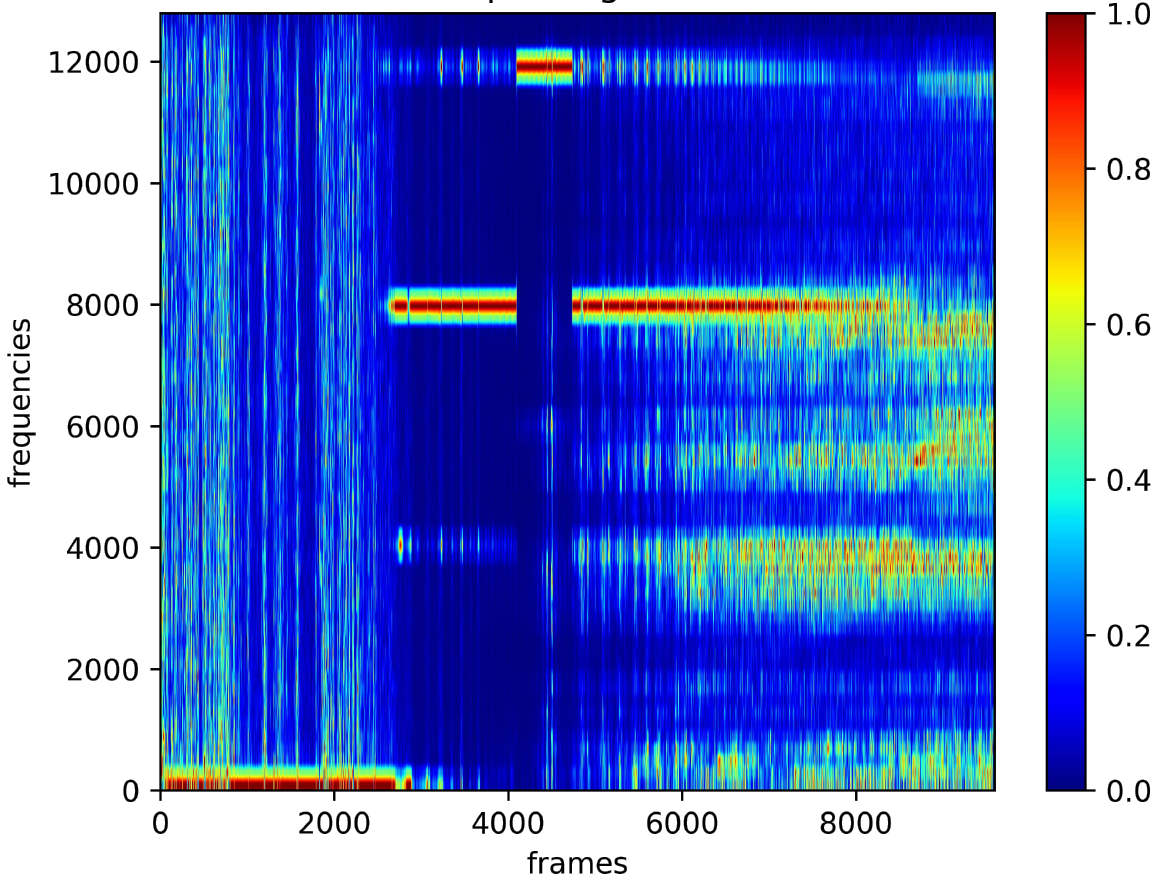}
		\subcaption{SQV\_acc\_DSTFT.}
		\label{1302}
	\end{minipage}
    \begin{minipage}{0.33\linewidth}
		\centering
		\includegraphics[width=1.0\linewidth]{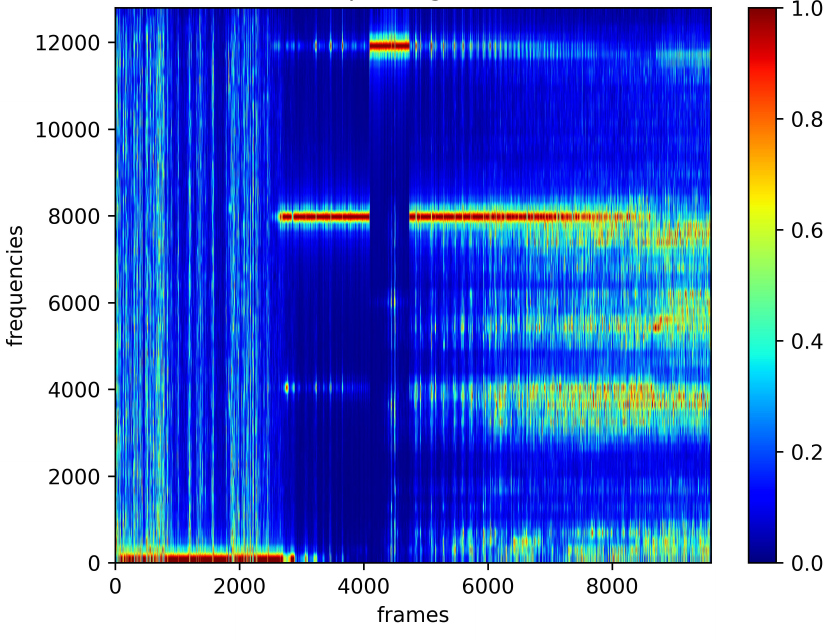}
		\subcaption{SQV\_acc\_MDSTFT.}
		\label{1303}
	\end{minipage}

    \begin{minipage}{0.33\linewidth}
		\centering
		\includegraphics[width=1.0\linewidth]{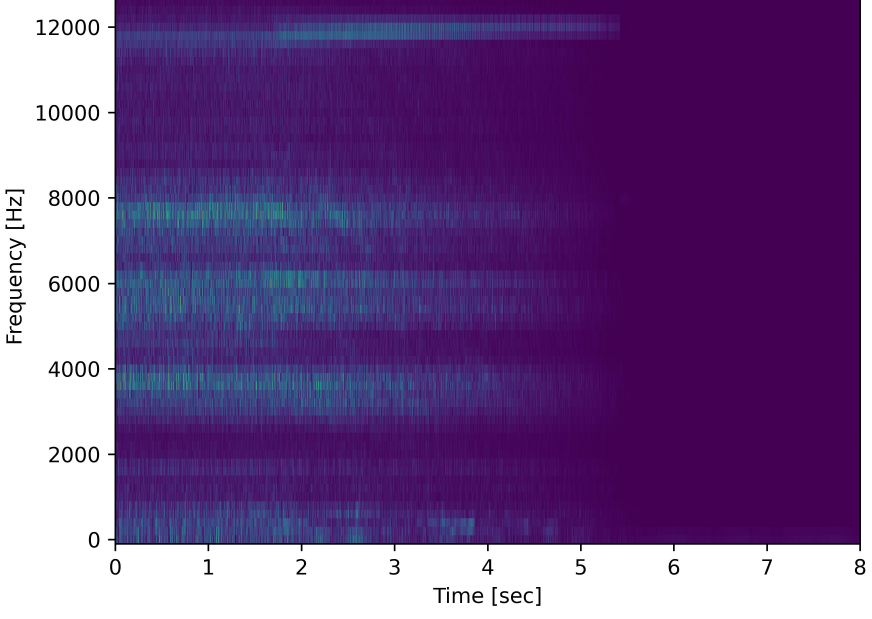}
		\subcaption{SQV\_dec\_STFT.}
		\label{1304}
	\end{minipage}
    \begin{minipage}{0.33\linewidth}
		\centering
		\includegraphics[width=1.0\linewidth]{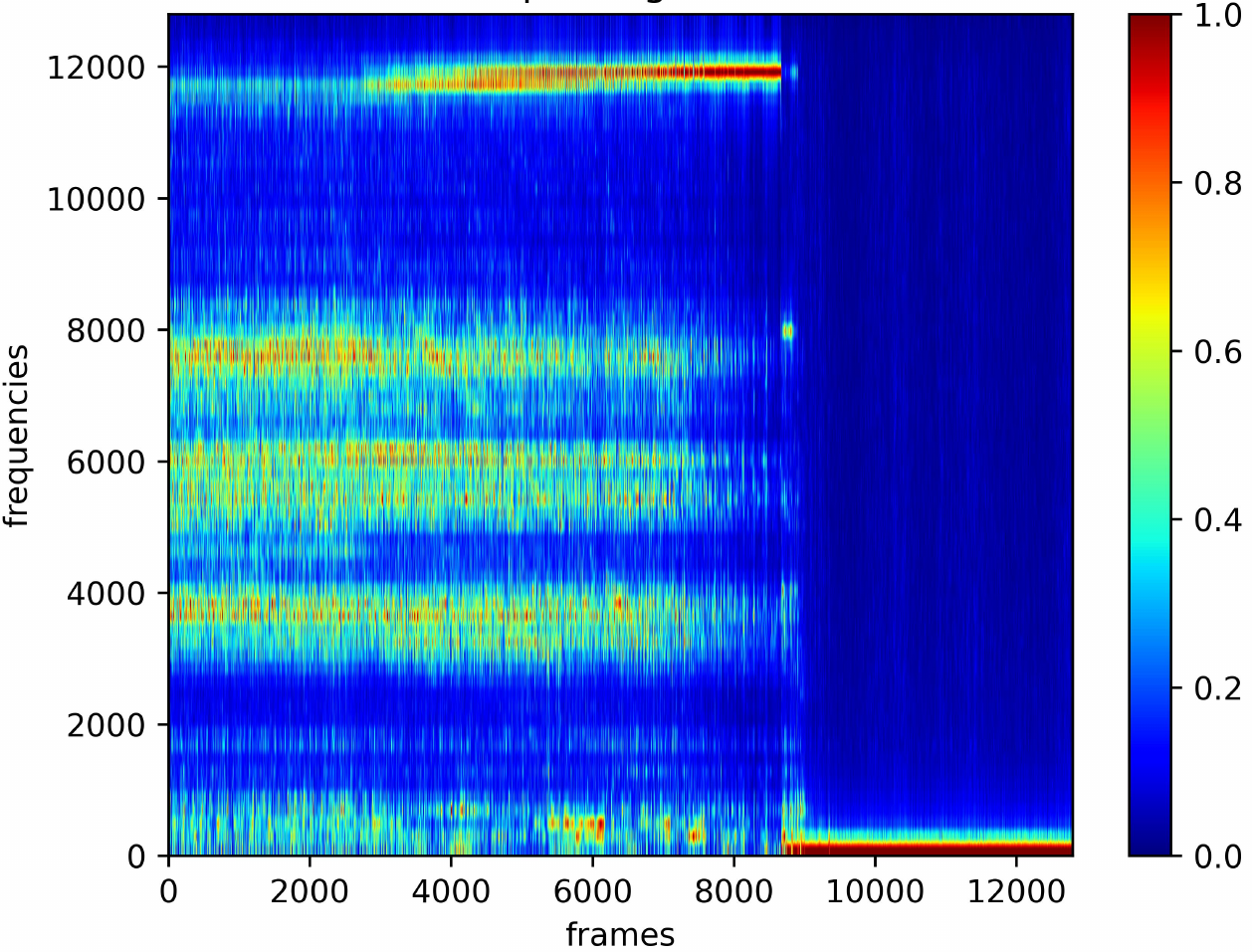}
		\subcaption{SQV\_dec\_DSTFT.}
		\label{1305}
	\end{minipage}
	\begin{minipage}{0.33\linewidth}
		\centering
		\includegraphics[width=1.0\linewidth]{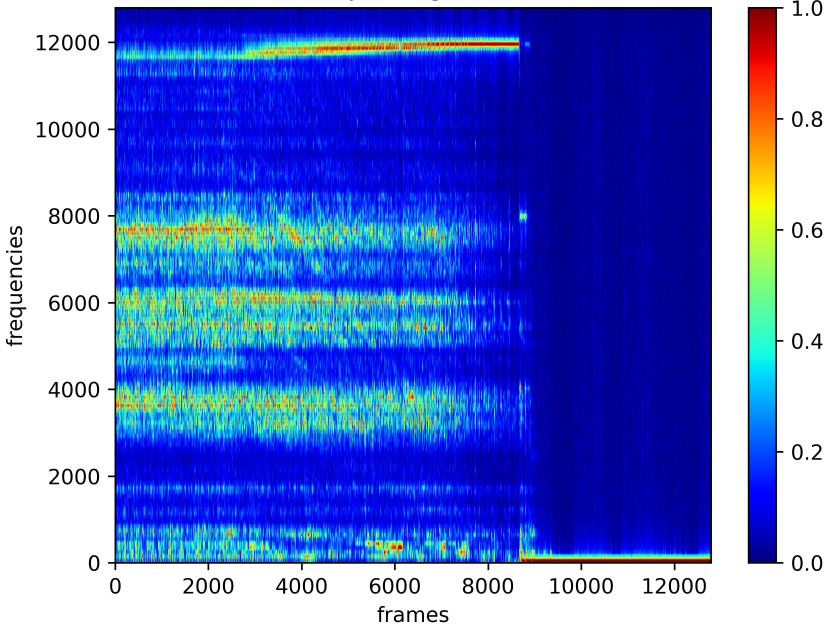}
		\subcaption{SQV\_dec\_MDSTFT.}
		\label{1306}
	\end{minipage}
  \caption{The spectrograms on SQV dataset with STFT, DSTFT, and MDSTFT.}
  \label{13}
\end{figure*}
As shown in \cref{13}, the  spectrograms for the outer ring fault is depicted. The energy focusability achieved by vanilla STFT is deficient and poor. While upon employing differentiable STFT, the energy focusability has significantly improved, which demonstrates superior aggregation capabilities. It can be seen that the main fault frequencies in the speed fluctuation data are mainly localized around 0Hz, 8000Hz, and 12000Hz during the acceleration process. The deceleration process tends to concentrate near 0Hz, 4000Hz, 6000Hz, 8000Hz, and 12000Hz. In summary, the proposed MDSTFT method exhibits the highest level of energy aggregation degree and more prominent fault characteristics.
\subsection{Quantitative perspective}
Furthermore, several methods are critically conducted quantitatively. Rényi entropy is extensively employed metric to evaluate energy aggregation, but it is incapable of back-propagate as it calculates the probability distribution density. Rényi entropy falls under the evaluation indicator, and BSQ pertains to the differentiable loss index. A decrease in both these indicators signifies higher spectrogram quality. From  \myyyref{tab05}, the Rényi entropy of STFT is the highest, indicating that the poorer energy focusability, the more diminished the quality of spectrograms. The proposed MDSTFT has the lowest Rényi entropy (1.398 and 1.7411) and BSQ (0.2515 and 0.3220), whose quantitative values are lower than DSTFT, highlighting the necessity of introducing the mask modulation mechanism.
\begin{table}[htbp]
  \centering
  \caption{Rényi entropy and BSQ of SQV dataset}
    \begin{tabular}{ccccc}
    \toprule
          &       & STFT  & DSTFT & MDSTFT \\
    \midrule
    \multirow{2}[4]{*}{SQV\_acc} & Rényi entropy & 1.4880 & 1.4100  & \cellcolor{gray!50}\textbf{1.3980} \\
\cmidrule{2-5}          & BSQ  &   \diagbox[dir=SW]{}     & 0.5548 & \cellcolor{gray!50}\textbf{0.2515} \\
    \midrule
    \multirow{2}[4]{*}{SQV\_dec} & Rényi entropy & 2.0830 & 1.9980 & \cellcolor{gray!50}\textbf{1.7410} \\
\cmidrule{2-5}          & BSQ  &    \diagbox[dir=SW]{}    & 0.5378 & \cellcolor{gray!50}\textbf{0.3220} \\
    \bottomrule
    \end{tabular}%
  \label{tab05}%
\end{table}%
\section{Conclusion}\label{section:06}
In this paper, we propose a physics-informed modulated differentiable STFT, which introduces modulation differentiable STFT to capture fault frequency information under time-varying speeds. Additionally, a loss function, termed as balanced spectrum, is devised to evaluate the quality of the generated spectrograms. By employing these two mechanisms, we have migrated the predicament of bearing transfer diagnosis in heavy haul freight trains under time-varying speed conditions, and have achieved the superior performance among analogous methods. For heavy haul freight equipment, and a future exploration of signal processing informed neural networks integrated physical constraints is a promising avenue.

Nonetheless, several issues worth further consideration. Initially, the proposed pyDSN needs to train two distinct models, because acceleration and deceleration are the distinct nature. Potential future research directions can focus on the cross-machine transfer diagnosis at time-varying full speeds via one model. Secondly, differentiable signal processing has broad application potential, such as differentiable wavelets and differentiable mode decomposition. Essentially, the proposed paradigm is an end-to-end strategy that combines signal processing prior-empowered modules and classifiers without conflicts, yet the process from feature to decision in the classifier remains a black box. In the future, using explainable structures similar to capsule networks as classifiers is also an interesting topic. Lastly, this study is limited to closed set transfer diagnosis, and the performance with differentiable STFT in partial domain or open set domain adaptation warrants additional scrutiny.

\section*{CRediT authorship contribution statement}
Chao He: Writing– original draft, Software, Validation, Visualization, Methodology, Investigation, Proofreading. Hongmei Shi: Conceptualization, Methodology, Writing - review \& editing, Supervision, Project administration, Funding acquisition. Jianbo Li: Supervision. ZuJun Yu: Supervision, Project administration.
\section*{Declaration of competing interest}
The authors declare that they have no known competing financial interests or personal relationships that could have appeared to influence the work reported in this paper.
\section*{Acknowledgments}
The authors are grateful for the supports of the National Natural Science Foundation of China (No. 52272429), and State Key Laboratory of Advanced Rail Autonomous Operation (Contract No. RAO2023ZZ003).

Su Li, for your companionship along the way, for the luck you have brought me, let's love each other forever, grow old together, and last for a long time.
\bibliographystyle{elsarticle-num-names}
\bibliography{cas-refs}

\begin{thebibliography}{82}
\providecommand{\natexlab}[1]{#1}
\providecommand{\url}[1]{\texttt{#1}}
\providecommand{\urlprefix}{URL }
\expandafter\ifx\csname urlstyle\endcsname\relax
  \providecommand{\doi}[1]{doi:\discretionary{}{}{}#1}\else
  \providecommand{\doi}[1]{doi:\discretionary{}{}{}\begingroup
  \urlstyle{rm}\url{#1}\endgroup}\fi
\providecommand{\bibinfo}[2]{#2}

\bibitem[{Mykhalkiv et~al.(2018)Mykhalkiv, Ravlyuk, Khodakivskyi, and
  Bereznyi}]{h45}
\bibinfo{author}{S.~Mykhalkiv}, \bibinfo{author}{V.~Ravlyuk},
  \bibinfo{author}{A.~Khodakivskyi}, \bibinfo{author}{V.~Bereznyi},
  \bibinfo{title}{Identification of Axle-Box Bearing Faults of Freight Cars
  Based on Minimum Entropy Deconvolution and Squared Envelope Spectra},
  \bibinfo{journal}{International Journal of Engineering and Technology(UAE)}
  \bibinfo{volume}{7} (\bibinfo{year}{2018}) \bibinfo{pages}{167--173},
  \bibinfo{note}{\doi{https://doi.org/10.14419/ijet.v7i4.3.19729}}.

\bibitem[{Wang et~al.(2021)Wang, Gao, Tang, Wang, Chen, Wang, Wang, and
  He}]{h46}
\bibinfo{author}{Q.~Wang}, \bibinfo{author}{T.~Gao}, \bibinfo{author}{H.~Tang},
  \bibinfo{author}{Y.~Wang}, \bibinfo{author}{Z.~Chen},
  \bibinfo{author}{J.~Wang}, \bibinfo{author}{P.~Wang},
  \bibinfo{author}{Q.~He}, \bibinfo{title}{A feature engineering framework for
  online fault diagnosis of freight train air brakes},
  \bibinfo{journal}{Measurement} \bibinfo{volume}{182} (\bibinfo{year}{2021})
  \bibinfo{pages}{109672},
  \bibinfo{note}{\doi{https://doi.org/10.1016/j.measurement.2021.109672}}.

\bibitem[{Si et~al.(2022{\natexlab{a}})Si, Shi, and Yang}]{h47}
\bibinfo{author}{J.~Si}, \bibinfo{author}{H.~Shi}, \bibinfo{author}{J.~Yang},
  \bibinfo{title}{Evaluation of Chinese freight train bearing condition based
  on spatiotemporal feature extraction}, \bibinfo{journal}{Proceedings of the
  Institution of Mechanical Engineers, Part F: Journal of Rail and Rapid
  Transit} \bibinfo{volume}{236}~(\bibinfo{number}{9})
  (\bibinfo{year}{2022}{\natexlab{a}}) \bibinfo{pages}{1047--1057},
  \bibinfo{note}{\doi{https://doi.org/10.1177/09544097211061944}}.

\bibitem[{Wu et~al.(2024)Wu, Li, Jia, An, Li, Antoni, and Xin}]{h69}
\bibinfo{author}{J.~Wu}, \bibinfo{author}{Y.~Li}, \bibinfo{author}{L.~Jia},
  \bibinfo{author}{G.~An}, \bibinfo{author}{Y.-F. Li},
  \bibinfo{author}{J.~Antoni}, \bibinfo{author}{G.~Xin},
  \bibinfo{title}{Semi-supervised fault diagnosis of wheelset bearings in
  high-speed trains using autocorrelation and improved flow Gaussian mixture
  model}, \bibinfo{journal}{Engineering Applications of Artificial
  Intelligence} \bibinfo{volume}{132} (\bibinfo{year}{2024})
  \bibinfo{pages}{107861},
  \bibinfo{note}{\doi{https://doi.org/10.1016/j.engappai.2024.107861}}.

\bibitem[{Liu et~al.(2021)Liu, Zhang, Liu, and Zhang}]{h44}
\bibinfo{author}{Z.~Liu}, \bibinfo{author}{M.~Zhang}, \bibinfo{author}{F.~Liu},
  \bibinfo{author}{B.~Zhang}, \bibinfo{title}{Multidimensional Feature Fusion
  and Ensemble Learning-Based Fault Diagnosis for the Braking System of
  Heavy-Haul Train}, \bibinfo{journal}{IEEE Transactions on Industrial
  Informatics} \bibinfo{volume}{17}~(\bibinfo{number}{1})
  (\bibinfo{year}{2021}) \bibinfo{pages}{41--51},
  \bibinfo{note}{\doi{https://doi.org/10.1109/TII.2020.2979467}}.

\bibitem[{Li et~al.(2021)Li, Yang, Tang, Wang, Li, and He}]{h48}
\bibinfo{author}{C.~Li}, \bibinfo{author}{K.~Yang}, \bibinfo{author}{H.~Tang},
  \bibinfo{author}{P.~Wang}, \bibinfo{author}{J.~Li}, \bibinfo{author}{Q.~He},
  \bibinfo{title}{Fault diagnosis for rolling bearings of a freight train under
  limited fault data: Few-shot learning method}, \bibinfo{journal}{Journal of
  Transportation Engineering, Part A: Systems}
  \bibinfo{volume}{147}~(\bibinfo{number}{8}) (\bibinfo{year}{2021})
  \bibinfo{pages}{04021041},
  \bibinfo{note}{\doi{https://doi.org/10.1061/JTEPBS.0000554}}.

\bibitem[{Liu et~al.(2023{\natexlab{a}})Liu, Cui, and Wang}]{h101}
\bibinfo{author}{D.~Liu}, \bibinfo{author}{L.~Cui}, \bibinfo{author}{H.~Wang},
  \bibinfo{title}{Rotating Machinery Fault Diagnosis Under Time-Varying Speeds:
  A Review}, \bibinfo{journal}{IEEE Sensors Journal}
  \bibinfo{volume}{23}~(\bibinfo{number}{24})
  (\bibinfo{year}{2023}{\natexlab{a}}) \bibinfo{pages}{29969--29990},
  \bibinfo{note}{\doi{https://doi.org/10.1109/JSEN.2023.3326112}}.

\bibitem[{Qian et~al.(2024)Qian, Luo, and Qin}]{h832}
\bibinfo{author}{Q.~Qian}, \bibinfo{author}{J.~Luo}, \bibinfo{author}{Y.~Qin},
  \bibinfo{title}{Adaptive Intermediate Class-Wise Distribution Alignment: A
  Universal Domain Adaptation and Generalization Method for Machine Fault
  Diagnosis}, \bibinfo{journal}{IEEE Transactions on Neural Networks and
  Learning Systems}  (\bibinfo{year}{2024})
  \bibinfo{pages}{1--15}\bibinfo{note}{\doi{https://doi.org/10.1109/TNNLS.2024.3376449}}.

\bibitem[{Han et~al.(2024)Han, Tian, Chung, and Wei}]{h72}
\bibinfo{author}{T.~Han}, \bibinfo{author}{J.~Tian},
  \bibinfo{author}{C.~Chung}, \bibinfo{author}{Y.~M. Wei},
  \bibinfo{title}{Challenges and opportunities for battery health estimation:
  Bridging laboratory research and real-world applications},
  \bibinfo{journal}{Journal of Energy Chemistry} \bibinfo{volume}{89}
  (\bibinfo{year}{2024}) \bibinfo{pages}{434--436},
  \bibinfo{note}{\doi{https://doi.org/10.1016/j.jechem.2023.10.032}}.

\bibitem[{Liao et~al.(2023{\natexlab{a}})Liao, Dong, Sun, Sun, Zhang, and
  Fan}]{h200}
\bibinfo{author}{J.-X. Liao}, \bibinfo{author}{H.-C. Dong},
  \bibinfo{author}{Z.-Q. Sun}, \bibinfo{author}{J.~Sun},
  \bibinfo{author}{S.~Zhang}, \bibinfo{author}{F.-L. Fan},
  \bibinfo{title}{Attention-embedded quadratic network (qttention) for
  effective and interpretable bearing fault diagnosis}, \bibinfo{journal}{IEEE
  Transactions on Instrumentation and Measurement} \bibinfo{volume}{72}
  (\bibinfo{year}{2023}{\natexlab{a}}) \bibinfo{pages}{1--13}.

\bibitem[{Yang et~al.(2024)Yang, Wang, Guo, Gong, and Shan}]{h102}
\bibinfo{author}{J.~Yang}, \bibinfo{author}{Z.~Wang}, \bibinfo{author}{Y.~Guo},
  \bibinfo{author}{T.~Gong}, \bibinfo{author}{Z.~Shan}, \bibinfo{title}{A Novel
  Noise-Aided Fault Feature Extraction Using Stochastic Resonance in a
  Nonlinear System and Its Application}, \bibinfo{journal}{IEEE Sensors
  Journal} \bibinfo{volume}{24}~(\bibinfo{number}{7}) (\bibinfo{year}{2024})
  \bibinfo{pages}{11856--11866},
  \bibinfo{note}{\doi{https://doi.org/10.1109/JSEN.2024.3365105}}.

\bibitem[{Liang et~al.(2022)Liang, Zhao, Lin, Jiao, and Ding}]{h104}
\bibinfo{author}{K.~Liang}, \bibinfo{author}{M.~Zhao},
  \bibinfo{author}{J.~Lin}, \bibinfo{author}{J.~Jiao},
  \bibinfo{author}{C.~Ding}, \bibinfo{title}{Toothwise Health Monitoring of
  Planetary Gearbox Under Time-Varying Speed Condition Based on Rotating
  Encoder Signal}, \bibinfo{journal}{IEEE Transactions on Industrial
  Electronics} \bibinfo{volume}{69}~(\bibinfo{number}{6})
  (\bibinfo{year}{2022}) \bibinfo{pages}{6267--6277},
  \bibinfo{note}{\doi{https://doi.org/10.1109/TIE.2021.3090713}}.

\bibitem[{Cheng et~al.(2023)Cheng, Yuan, Lu, Wang, Ding, and Gong}]{h105}
\bibinfo{author}{X.~Cheng}, \bibinfo{author}{L.~Yuan}, \bibinfo{author}{Y.~Lu},
  \bibinfo{author}{Y.~Wang}, \bibinfo{author}{N.~Ding},
  \bibinfo{author}{Y.~Gong}, \bibinfo{title}{Fault identification of rolling
  bearings under linear varying speed based on the slope features of
  time--frequency ridges}, \bibinfo{journal}{Mechanical Systems and Signal
  Processing} \bibinfo{volume}{205} (\bibinfo{year}{2023})
  \bibinfo{pages}{110834},
  \bibinfo{note}{\doi{https://doi.org/10.1016/j.ymssp.2023.110834}}.

\bibitem[{Li et~al.(2022)Li, Shao, Lu, Xiang, and Cai}]{h11}
\bibinfo{author}{X.~Li}, \bibinfo{author}{H.~Shao}, \bibinfo{author}{S.~Lu},
  \bibinfo{author}{J.~Xiang}, \bibinfo{author}{B.~Cai}, \bibinfo{title}{Highly
  Efficient Fault Diagnosis of Rotating Machinery Under Time-Varying Speeds
  Using LSISMM and Small Infrared Thermal Images}, \bibinfo{journal}{IEEE
  Transactions on Systems, Man, and Cybernetics: Systems}
  \bibinfo{volume}{52}~(\bibinfo{number}{12}) (\bibinfo{year}{2022})
  \bibinfo{pages}{7328--7340},
  \bibinfo{note}{\doi{https://doi.org/10.1109/TSMC.2022.3151185}}.

\bibitem[{Shao et~al.(2023)Shao, Li, Cai, Wan, Xiao, and Yan}]{h111}
\bibinfo{author}{H.~Shao}, \bibinfo{author}{W.~Li}, \bibinfo{author}{B.~Cai},
  \bibinfo{author}{J.~Wan}, \bibinfo{author}{Y.~Xiao},
  \bibinfo{author}{S.~Yan}, \bibinfo{title}{Dual-Threshold Attention-Guided GAN
  and Limited Infrared Thermal Images for Rotating Machinery Fault Diagnosis
  Under Speed Fluctuation}, \bibinfo{journal}{IEEE Transactions on Industrial
  Informatics} \bibinfo{volume}{19}~(\bibinfo{number}{9})
  (\bibinfo{year}{2023}) \bibinfo{pages}{9933--9942},
  \bibinfo{note}{\doi{https://doi.org/10.1109/TII.2022.3232766}}.

\bibitem[{Xu et~al.(2023{\natexlab{a}})Xu, Tao, Li, and Zhong}]{h12}
\bibinfo{author}{Y.~Xu}, \bibinfo{author}{H.~Tao}, \bibinfo{author}{W.~Li},
  \bibinfo{author}{Y.~Zhong}, \bibinfo{title}{CapsFormer: A Novel Bearing
  Intelligent Fault Diagnosis Framework With Negligible Speed Change Under
  Small-Sample Conditions}, \bibinfo{journal}{IEEE Transactions on
  Instrumentation and Measurement} \bibinfo{volume}{72}
  (\bibinfo{year}{2023}{\natexlab{a}}) \bibinfo{pages}{1--11},
  \bibinfo{note}{\doi{https://doi.org/10.1109/TIM.2023.3318693}}.

\bibitem[{Rao et~al.(2023{\natexlab{a}})Rao, Zuo, and Tian}]{h13}
\bibinfo{author}{M.~Rao}, \bibinfo{author}{M.~J. Zuo},
  \bibinfo{author}{Z.~Tian}, \bibinfo{title}{A speed normalized autoencoder for
  rotating machinery fault detection under varying speed conditions},
  \bibinfo{journal}{Mechanical Systems and Signal Processing}
  \bibinfo{volume}{189} (\bibinfo{year}{2023}{\natexlab{a}})
  \bibinfo{pages}{110109},
  \bibinfo{note}{\doi{https://doi.org/10.1016/j.ymssp.2023.110109}}.

\bibitem[{Rao et~al.(2023{\natexlab{b}})Rao, Zuo, and Tian}]{h14}
\bibinfo{author}{M.~Rao}, \bibinfo{author}{M.~J. Zuo},
  \bibinfo{author}{Z.~Tian}, \bibinfo{title}{Speed adaptive gate: A novel
  auxiliary branch for enhancing deep learning-based rotating machinery fault
  classification under varying speed conditions},
  \bibinfo{journal}{Measurement} \bibinfo{volume}{217}
  (\bibinfo{year}{2023}{\natexlab{b}}) \bibinfo{pages}{113016},
  \bibinfo{note}{\doi{https://doi.org/10.1016/j.measurement.2023.113016}}.

\bibitem[{Shi et~al.(2023{\natexlab{a}})Shi, Chen, Zi, and Chen}]{h16}
\bibinfo{author}{Z.~Shi}, \bibinfo{author}{J.~Chen}, \bibinfo{author}{Y.~Zi},
  \bibinfo{author}{Z.~Chen}, \bibinfo{title}{DecouplingNet: A Stable Knowledge
  Distillation Decoupling Net for Fault Detection of Rotating Machines Under
  Varying Speeds}, \bibinfo{journal}{IEEE Transactions on Neural Networks and
  Learning Systems}  (\bibinfo{year}{2023}{\natexlab{a}})
  \bibinfo{pages}{1--15}\bibinfo{note}{\doi{https://doi.org/10.1109/TNNLS.2023.3258748}}.

\bibitem[{Li et~al.(2023)Li, Wu, Li, and Cheng}]{h15}
\bibinfo{author}{R.~Li}, \bibinfo{author}{J.~Wu}, \bibinfo{author}{Y.~Li},
  \bibinfo{author}{Y.~Cheng}, \bibinfo{title}{PeriodNet: Noise-Robust Fault
  Diagnosis Method Under Varying Speed Conditions}, \bibinfo{journal}{IEEE
  Transactions on Neural Networks and Learning Systems}  (\bibinfo{year}{2023})
  \bibinfo{pages}{1--15}\bibinfo{note}{\doi{https://doi.org/10.1109/TNNLS.2023.3274290}}.

\bibitem[{Yan et~al.(2023)Yan, She, and Xu}]{h17}
\bibinfo{author}{X.~Yan}, \bibinfo{author}{D.~She}, \bibinfo{author}{Y.~Xu},
  \bibinfo{title}{Deep order-wavelet convolutional variational autoencoder for
  fault identification of rolling bearing under fluctuating speed conditions},
  \bibinfo{journal}{Expert Systems with Applications} \bibinfo{volume}{216}
  (\bibinfo{year}{2023}) \bibinfo{pages}{119479},
  \bibinfo{note}{\doi{https://doi.org/10.1016/j.eswa.2022.119479}}.

\bibitem[{Chang et~al.(2022)Chang, Chen, Chen, Liu, and Zhou}]{h18}
\bibinfo{author}{Y.~Chang}, \bibinfo{author}{J.~Chen},
  \bibinfo{author}{Q.~Chen}, \bibinfo{author}{S.~Liu},
  \bibinfo{author}{Z.~Zhou}, \bibinfo{title}{CFs-focused intelligent diagnosis
  scheme via alternative kernels networks with soft squeeze-and-excitation
  attention for fast-precise fault detection under slow \& sharp speed
  variations}, \bibinfo{journal}{Knowledge-Based Systems} \bibinfo{volume}{239}
  (\bibinfo{year}{2022}) \bibinfo{pages}{108026},
  \bibinfo{note}{\doi{https://doi.org/10.1016/j.knosys.2021.108026}}.

\bibitem[{Yu et~al.(2019)Yu, Xu, and Liu}]{h19}
\bibinfo{author}{J.~Yu}, \bibinfo{author}{Y.~Xu}, \bibinfo{author}{K.~Liu},
  \bibinfo{title}{Planetary gear fault diagnosis using stacked denoising
  autoencoder and gated recurrent unit neural network under noisy environment
  and time-varying rotational speed conditions}, \bibinfo{journal}{Measurement
  Science and Technology} \bibinfo{volume}{30}~(\bibinfo{number}{9})
  (\bibinfo{year}{2019}) \bibinfo{pages}{095003},
  \bibinfo{note}{\doi{https://doi.org/10.1088/1361-6501/ab1da0}}.

\bibitem[{An et~al.(2020)An, Li, Wang, and Jiang}]{h20}
\bibinfo{author}{Z.~An}, \bibinfo{author}{S.~Li}, \bibinfo{author}{J.~Wang},
  \bibinfo{author}{X.~Jiang}, \bibinfo{title}{A novel bearing intelligent fault
  diagnosis framework under time-varying working conditions using recurrent
  neural network}, \bibinfo{journal}{ISA transactions} \bibinfo{volume}{100}
  (\bibinfo{year}{2020}) \bibinfo{pages}{155--170},
  \bibinfo{note}{\doi{https://doi.org/10.1016/j.isatra.2019.11.010}}.

\bibitem[{Shi et~al.(2022)Shi, Liu, Chen, Zi, and Zhou}]{h21}
\bibinfo{author}{Z.~Shi}, \bibinfo{author}{X.~Liu}, \bibinfo{author}{J.~Chen},
  \bibinfo{author}{Y.~Zi}, \bibinfo{author}{Z.~Zhou}, \bibinfo{title}{A
  multi-branch redundant adversarial net for intelligent fault diagnosis of
  multiple components under drastically variable speeds}, \bibinfo{journal}{ISA
  transactions} \bibinfo{volume}{129} (\bibinfo{year}{2022})
  \bibinfo{pages}{540--554},
  \bibinfo{note}{\doi{https://doi.org/10.1016/j.isatra.2022.01.011}}.

\bibitem[{Pu et~al.(2023)Pu, Zhang, and An}]{h211}
\bibinfo{author}{H.~Pu}, \bibinfo{author}{K.~Zhang}, \bibinfo{author}{Y.~An},
  \bibinfo{title}{Restricted Sparse Networks for Rolling Bearing Fault
  Diagnosis}, \bibinfo{journal}{IEEE Transactions on Industrial Informatics}
  \bibinfo{volume}{19}~(\bibinfo{number}{11}) (\bibinfo{year}{2023})
  \bibinfo{pages}{11139--11149},
  \bibinfo{note}{\doi{https://doi.org/10.1109/TII.2023.3243929}}.

\bibitem[{Guo et~al.(2024)Guo, He, and Gu}]{h71}
\bibinfo{author}{J.~Guo}, \bibinfo{author}{Q.~He}, \bibinfo{author}{F.~Gu},
  \bibinfo{title}{DNOCNet: A Novel End-to-End Network for Induction Motor Drive
  Systems Fault Diagnosis Under Speed Fluctuation Condition},
  \bibinfo{journal}{IEEE Transactions on Industrial Informatics}
  (\bibinfo{year}{2024})
  \bibinfo{pages}{1--10}\bibinfo{note}{\doi{https://doi.org/10.1109/TII.2024.3369239}}.

\bibitem[{Yan et~al.(2022{\natexlab{a}})Yan, Shao, Xiao, Zhou, Xu, and
  Wan}]{h22}
\bibinfo{author}{S.~Yan}, \bibinfo{author}{H.~Shao}, \bibinfo{author}{Y.~Xiao},
  \bibinfo{author}{J.~Zhou}, \bibinfo{author}{Y.~Xu}, \bibinfo{author}{J.~Wan},
  \bibinfo{title}{Semi-supervised fault diagnosis of machinery using LPS-DGAT
  under speed fluctuation and extremely low labeled rates},
  \bibinfo{journal}{Advanced Engineering Informatics} \bibinfo{volume}{53}
  (\bibinfo{year}{2022}{\natexlab{a}}) \bibinfo{pages}{101648},
  \bibinfo{note}{\doi{https://doi.org/10.1016/j.aei.2022.101648}}.

\bibitem[{Dong et~al.(2024)Dong, Jiang, Yao, Mu, and Yang}]{h38}
\bibinfo{author}{Y.~Dong}, \bibinfo{author}{H.~Jiang},
  \bibinfo{author}{R.~Yao}, \bibinfo{author}{M.~Mu}, \bibinfo{author}{Q.~Yang},
  \bibinfo{title}{Rolling bearing intelligent fault diagnosis towards variable
  speed and imbalanced samples using multiscale dynamic supervised contrast
  learning}, \bibinfo{journal}{Reliability Engineering \& System Safety}
  \bibinfo{volume}{243} (\bibinfo{year}{2024}) \bibinfo{pages}{109805},
  \bibinfo{note}{\doi{https://doi.org/10.1016/j.ress.2023.109805}}.

\bibitem[{Liu et~al.(2023{\natexlab{b}})Liu, Yan, Liu, Wang, Huang, and
  Wu}]{h834}
\bibinfo{author}{B.~Liu}, \bibinfo{author}{C.~Yan}, \bibinfo{author}{Y.~Liu},
  \bibinfo{author}{Z.~Wang}, \bibinfo{author}{Y.~Huang},
  \bibinfo{author}{L.~Wu}, \bibinfo{title}{Multiscale Residual Antinoise
  Network via Interpretable Dynamic Recalibration Mechanism for Rolling Bearing
  Fault Diagnosis With Few Samples}, \bibinfo{journal}{IEEE Sensors Journal}
  \bibinfo{volume}{23}~(\bibinfo{number}{24})
  (\bibinfo{year}{2023}{\natexlab{b}}) \bibinfo{pages}{31425--31439},
  \bibinfo{note}{\doi{https://doi.org/10.1109/JSEN.2023.3328007}}.

\bibitem[{Meng et~al.(2023)Meng, He, Cao, Li, Cao, Fan, Zhu, and Fan}]{h836}
\bibinfo{author}{Z.~Meng}, \bibinfo{author}{H.~He}, \bibinfo{author}{W.~Cao},
  \bibinfo{author}{J.~Li}, \bibinfo{author}{L.~Cao}, \bibinfo{author}{J.~Fan},
  \bibinfo{author}{M.~Zhu}, \bibinfo{author}{F.~Fan}, \bibinfo{title}{A novel
  generation network using feature fusion and guided adversarial learning for
  fault diagnosis of rotating machinery}, \bibinfo{journal}{Expert Systems with
  Applications} \bibinfo{volume}{234} (\bibinfo{year}{2023})
  \bibinfo{pages}{121058},
  \bibinfo{note}{\doi{https://doi.org/10.1016/j.asoc.2023.110669}}.

\bibitem[{He et~al.(2023)He, Shi, and Li}]{h10}
\bibinfo{author}{C.~He}, \bibinfo{author}{H.~Shi}, \bibinfo{author}{J.~Li},
  \bibinfo{title}{IDSN: A one-stage interpretable and differentiable STFT
  domain adaptation network for traction motor of high-speed trains
  cross-machine diagnosis}, \bibinfo{journal}{Mechanical Systems and Signal
  Processing} \bibinfo{volume}{205} (\bibinfo{year}{2023})
  \bibinfo{pages}{110846},
  \bibinfo{note}{\doi{https://doi.org/10.1016/j.ymssp.2023.110846}}.

\bibitem[{Han et~al.(2023)Han, Xie, and Pei}]{h74}
\bibinfo{author}{T.~Han}, \bibinfo{author}{W.~Xie}, \bibinfo{author}{Z.~Pei},
  \bibinfo{title}{Semi-supervised adversarial discriminative learning approach
  for intelligent fault diagnosis of wind turbine},
  \bibinfo{journal}{Information Sciences} \bibinfo{volume}{648}
  (\bibinfo{year}{2023}) \bibinfo{pages}{119496},
  \bibinfo{note}{\doi{https://doi.org/10.1016/j.ins.2023.119496}}.

\bibitem[{Li et~al.(2024)Li, Wang, Yao, Li, and Gao}]{h833}
\bibinfo{author}{X.~Li}, \bibinfo{author}{Y.~Wang}, \bibinfo{author}{J.~Yao},
  \bibinfo{author}{M.~Li}, \bibinfo{author}{Z.~Gao},
  \bibinfo{title}{Multi-sensor fusion fault diagnosis method of wind turbine
  bearing based on adaptive convergent viewable neural networks},
  \bibinfo{journal}{Reliability Engineering \& System Safety}
  \bibinfo{volume}{245} (\bibinfo{year}{2024}) \bibinfo{pages}{109980},
  \bibinfo{note}{\doi{https://doi.org/10.1016/j.ress.2024.109980}}.

\bibitem[{Sun et~al.(2023)Sun, Meng, Guan, Liu, Cao, and Fan}]{h835}
\bibinfo{author}{D.~Sun}, \bibinfo{author}{Z.~Meng}, \bibinfo{author}{Y.~Guan},
  \bibinfo{author}{J.~Liu}, \bibinfo{author}{W.~Cao}, \bibinfo{author}{F.~Fan},
  \bibinfo{title}{Intelligent fault diagnosis scheme for rolling bearing based
  on domain adaptation in one dimensional feature matching},
  \bibinfo{journal}{Applied Soft Computing} \bibinfo{volume}{146}
  (\bibinfo{year}{2023}) \bibinfo{pages}{110669},
  \bibinfo{note}{\doi{https://doi.org/10.1016/j.eswa.2023.121058}}.

\bibitem[{Liang et~al.(2023)Liang, Wang, Jiang, Li, and Zhang}]{h24}
\bibinfo{author}{P.~Liang}, \bibinfo{author}{B.~Wang},
  \bibinfo{author}{G.~Jiang}, \bibinfo{author}{N.~Li},
  \bibinfo{author}{L.~Zhang}, \bibinfo{title}{Unsupervised fault diagnosis of
  wind turbine bearing via a deep residual deformable convolution network based
  on subdomain adaptation under time-varying speeds},
  \bibinfo{journal}{Engineering Applications of Artificial Intelligence}
  \bibinfo{volume}{118} (\bibinfo{year}{2023}) \bibinfo{pages}{105656},
  \bibinfo{note}{\doi{https://doi.org/10.1016/j.engappai.2022.105656}}.

\bibitem[{Liang et~al.(2024)Liang, Xu, Shuai, Yuan, Wang, and Zhang}]{h25}
\bibinfo{author}{P.~Liang}, \bibinfo{author}{L.~Xu},
  \bibinfo{author}{H.~Shuai}, \bibinfo{author}{X.~Yuan},
  \bibinfo{author}{B.~Wang}, \bibinfo{author}{L.~Zhang},
  \bibinfo{title}{Semisupervised Subdomain Adaptation Graph Convolutional
  Network for Fault Transfer Diagnosis of Rotating Machinery Under Time-Varying
  Speeds}, \bibinfo{journal}{IEEE/ASME Transactions on Mechatronics}
  \bibinfo{volume}{29}~(\bibinfo{number}{1}) (\bibinfo{year}{2024})
  \bibinfo{pages}{730--741},
  \bibinfo{note}{\doi{https://doi.org/10.1109/TMECH.2023.3292969}}.

\bibitem[{Cao et~al.(2022)Cao, Shao, Zhong, Deng, Yang, and Xuan}]{h30}
\bibinfo{author}{H.~Cao}, \bibinfo{author}{H.~Shao},
  \bibinfo{author}{X.~Zhong}, \bibinfo{author}{Q.~Deng},
  \bibinfo{author}{X.~Yang}, \bibinfo{author}{J.~Xuan},
  \bibinfo{title}{Unsupervised domain-share CNN for machine fault transfer
  diagnosis from steady speeds to time-varying speeds},
  \bibinfo{journal}{Journal of Manufacturing Systems} \bibinfo{volume}{62}
  (\bibinfo{year}{2022}) \bibinfo{pages}{186--198},
  \bibinfo{note}{\doi{https://doi.org/10.1016/j.jmsy.2021.11.016}}.

\bibitem[{Si et~al.(2022{\natexlab{b}})Si, Shi, Han, Chen, and Zheng}]{h28}
\bibinfo{author}{J.~Si}, \bibinfo{author}{H.~Shi}, \bibinfo{author}{T.~Han},
  \bibinfo{author}{J.~Chen}, \bibinfo{author}{C.~Zheng}, \bibinfo{title}{Learn
  Generalized Features Via Multi-Source Domain Adaptation: Intelligent
  Diagnosis Under Variable/Constant Machine Conditions}, \bibinfo{journal}{IEEE
  Sensors Journal} \bibinfo{volume}{22}~(\bibinfo{number}{1})
  (\bibinfo{year}{2022}{\natexlab{b}}) \bibinfo{pages}{510--519},
  \bibinfo{note}{\doi{https://doi.org/10.1109/JSEN.2021.3126864}}.

\bibitem[{Zhao et~al.(2023)Zhao, Zhu, Yao, Deng, Cao, Ding, Jia, and Shao}]{h6}
\bibinfo{author}{X.~Zhao}, \bibinfo{author}{X.~Zhu}, \bibinfo{author}{J.~Yao},
  \bibinfo{author}{W.~Deng}, \bibinfo{author}{Y.~Cao},
  \bibinfo{author}{P.~Ding}, \bibinfo{author}{M.~Jia},
  \bibinfo{author}{H.~Shao}, \bibinfo{title}{Intelligent Health Assessment of
  Aviation Bearing Based on Deep Transfer Graph Convolutional Networks under
  Large Speed Fluctuations}, \bibinfo{journal}{Sensors}
  \bibinfo{volume}{23}~(\bibinfo{number}{9}) (\bibinfo{year}{2023})
  \bibinfo{pages}{4379},
  \bibinfo{note}{\doi{https://doi.org/10.3390/s23094379}}.

\bibitem[{Shi et~al.(2023{\natexlab{b}})Shi, Chen, Zhang, Zi, Li, and
  Chen}]{h27}
\bibinfo{author}{Z.~Shi}, \bibinfo{author}{J.~Chen},
  \bibinfo{author}{X.~Zhang}, \bibinfo{author}{Y.~Zi}, \bibinfo{author}{C.~Li},
  \bibinfo{author}{J.~Chen}, \bibinfo{title}{A reliable feature-assisted
  contrastive generalization net for intelligent fault diagnosis under unseen
  machines and working conditions}, \bibinfo{journal}{Mechanical Systems and
  Signal Processing} \bibinfo{volume}{188} (\bibinfo{year}{2023}{\natexlab{b}})
  \bibinfo{pages}{110011},
  \bibinfo{note}{\doi{https://doi.org/10.1016/j.ymssp.2022.110011}}.

\bibitem[{Zhou et~al.(2022)Zhou, Huang, Wen, Dong, Lei, Zhang, and Chen}]{h29}
\bibinfo{author}{H.~Zhou}, \bibinfo{author}{X.~Huang},
  \bibinfo{author}{G.~Wen}, \bibinfo{author}{S.~Dong},
  \bibinfo{author}{Z.~Lei}, \bibinfo{author}{P.~Zhang},
  \bibinfo{author}{X.~Chen}, \bibinfo{title}{Convolution enabled transformer
  via random contrastive regularization for rotating machinery diagnosis under
  time-varying working conditions}, \bibinfo{journal}{Mechanical Systems and
  Signal Processing} \bibinfo{volume}{173} (\bibinfo{year}{2022})
  \bibinfo{pages}{109050},
  \bibinfo{note}{\doi{https://doi.org/10.1016/j.ymssp.2022.109050}}.

\bibitem[{Gao et~al.(2023)Gao, Huang, Zhu, Zhu, Yan, Ren, and Soares}]{h26}
\bibinfo{author}{D.~Gao}, \bibinfo{author}{K.~Huang}, \bibinfo{author}{Y.~Zhu},
  \bibinfo{author}{L.~Zhu}, \bibinfo{author}{K.~Yan}, \bibinfo{author}{Z.~Ren},
  \bibinfo{author}{C.~G. Soares}, \bibinfo{title}{Semi-supervised small sample
  fault diagnosis under a wide range of speed variation conditions based on
  uncertainty analysis}, \bibinfo{journal}{Reliability Engineering \& System
  Safety} \bibinfo{volume}{242} (\bibinfo{year}{2023}) \bibinfo{pages}{109746},
  \bibinfo{note}{\doi{https://doi.org/10.1016/j.ress.2023.109746}}.

\bibitem[{Xu et~al.(2023{\natexlab{b}})Xu, Tang, Pang, and Qi}]{h32}
\bibinfo{author}{Z.~Xu}, \bibinfo{author}{G.~Tang}, \bibinfo{author}{B.~Pang},
  \bibinfo{author}{X.~Qi}, \bibinfo{title}{Rolling bearing fault diagnosis
  under time-varying speeds based on time-characteristic order spectrum and
  multi-scale domain adaptation network}, \bibinfo{journal}{Measurement Science
  and Technology} \bibinfo{volume}{34}~(\bibinfo{number}{12})
  (\bibinfo{year}{2023}{\natexlab{b}}) \bibinfo{pages}{125118},
  \bibinfo{note}{\doi{https://doi.org/10.1088/1361-6501/acf332}}.

\bibitem[{Lu et~al.(2023)Lu, Tong, Feng, and Wan}]{h33}
\bibinfo{author}{F.~Lu}, \bibinfo{author}{Q.~Tong}, \bibinfo{author}{Z.~Feng},
  \bibinfo{author}{Q.~Wan}, \bibinfo{title}{Unbalanced Bearing Fault Diagnosis
  Under Various Speeds Based on Spectrum Alignment and Deep Transfer
  Convolution Neural Network}, \bibinfo{journal}{IEEE Transactions on
  Industrial Informatics} \bibinfo{volume}{19}~(\bibinfo{number}{7})
  (\bibinfo{year}{2023}) \bibinfo{pages}{8295--8306},
  \bibinfo{note}{\doi{https://doi.org/10.1109/TII.2022.3217541}}.

\bibitem[{Pang et~al.(2024)Pang, Liu, Sun, Xu, and Hao}]{h42}
\bibinfo{author}{B.~Pang}, \bibinfo{author}{Q.~Liu}, \bibinfo{author}{Z.~Sun},
  \bibinfo{author}{Z.~Xu}, \bibinfo{author}{Z.~Hao},
  \bibinfo{title}{Time-frequency supervised contrastive learning via
  pseudo-labeling: An unsupervised domain adaptation network for rolling
  bearing fault diagnosis under time-varying speeds},
  \bibinfo{journal}{Advanced Engineering Informatics} \bibinfo{volume}{59}
  (\bibinfo{year}{2024}) \bibinfo{pages}{102304},
  \bibinfo{note}{\doi{https://doi.org/10.1016/j.aei.2023.102304}}.

\bibitem[{Luo et~al.(2024)Luo, Shao, Lin, and Liu}]{h68}
\bibinfo{author}{J.~Luo}, \bibinfo{author}{H.~Shao}, \bibinfo{author}{J.~Lin},
  \bibinfo{author}{B.~Liu}, \bibinfo{title}{Meta-learning with elastic
  prototypical network for fault transfer diagnosis of bearings under unstable
  speeds}, \bibinfo{journal}{Reliability Engineering \& System Safety}
  \bibinfo{volume}{245} (\bibinfo{year}{2024}) \bibinfo{pages}{110001},
  \bibinfo{note}{\doi{https://doi.org/10.1016/j.ress.2024.110001}}.

\bibitem[{Wang et~al.(2023)Wang, Sun, and Wang}]{h34}
\bibinfo{author}{C.~Wang}, \bibinfo{author}{Y.~Sun}, \bibinfo{author}{X.~Wang},
  \bibinfo{title}{Image deep learning in fault diagnosis of mechanical
  equipment}, \bibinfo{journal}{Journal of Intelligent Manufacturing}
  (\bibinfo{year}{2023})
  \bibinfo{pages}{1--41}\bibinfo{note}{\doi{https://doi.org/10.1007/s10845-023-02176-3}}.

\bibitem[{Zhao et~al.(2021)Zhao, Subramani, and Smaragdis}]{h35}
\bibinfo{author}{A.~Zhao}, \bibinfo{author}{K.~Subramani},
  \bibinfo{author}{P.~Smaragdis}, \bibinfo{title}{Optimizing Short-Time Fourier
  Transform Parameters via Gradient Descent}, in: \bibinfo{booktitle}{{IEEE}
  International Conference on Acoustics, Speech and Signal Processing, {ICASSP}
  2021, Toronto, ON, Canada, June 6-11, 2021}, \bibinfo{pages}{736--740},
  \bibinfo{note}{\doi{https://doi.org/10.1109/ICASSP39728.2021.9413704}},
  \bibinfo{year}{2021}.

\bibitem[{Sun and Wang(2023)}]{h66}
\bibinfo{author}{Y.~Sun}, \bibinfo{author}{W.~Wang}, \bibinfo{title}{Role of
  image feature enhancement in intelligent fault diagnosis for mechanical
  equipment: A review}, \bibinfo{journal}{Engineering Failure Analysis}
  \bibinfo{volume}{156} (\bibinfo{year}{2023}) \bibinfo{pages}{107815},
  \bibinfo{note}{\doi{https://doi.org/10.1016/j.engfailanal.2023.107815}}.

\bibitem[{Xu et~al.(2023{\natexlab{c}})Xu, Kohtz, Boakye, Gardoni, and
  Wang}]{h75}
\bibinfo{author}{Y.~Xu}, \bibinfo{author}{S.~Kohtz},
  \bibinfo{author}{J.~Boakye}, \bibinfo{author}{P.~Gardoni},
  \bibinfo{author}{P.~Wang}, \bibinfo{title}{Physics-informed machine learning
  for reliability and systems safety applications: State of the art and
  challenges}, \bibinfo{journal}{Reliability Engineering \& System Safety}
  \bibinfo{volume}{230} (\bibinfo{year}{2023}{\natexlab{c}})
  \bibinfo{pages}{108900},
  \bibinfo{note}{\doi{https://doi.org/10.1016/j.ress.2022.108900}}.

\bibitem[{Yan et~al.(2024)Yan, Shang, Wang, Xu, Zhao, Wang, and Chen}]{h76}
\bibinfo{author}{R.~Yan}, \bibinfo{author}{Z.~Shang},
  \bibinfo{author}{Z.~Wang}, \bibinfo{author}{W.~Xu},
  \bibinfo{author}{Z.~Zhao}, \bibinfo{author}{S.~Wang},
  \bibinfo{author}{X.~Chen}, \bibinfo{title}{Challenges and Opportunities of
  XAI in Industrial Intelligent Diagnosis: Priori-empowered},
  \bibinfo{journal}{Journal of Mechanical Engineering} \bibinfo{volume}{60}
  (\bibinfo{year}{2024}) \bibinfo{pages}{1--20},
  \bibinfo{note}{\doi{http://kns.cnki.net/kcms/detail/11.2187.TH.20240220.1637.010.html.}}

\bibitem[{Liao et~al.(2023{\natexlab{b}})Liao, Dong, Luo, Sun, and Zhang}]{h77}
\bibinfo{author}{J.-X. Liao}, \bibinfo{author}{H.-C. Dong},
  \bibinfo{author}{L.~Luo}, \bibinfo{author}{J.~Sun},
  \bibinfo{author}{S.~Zhang}, \bibinfo{title}{Multi-task neural network blind
  deconvolution and its application to bearing fault feature extraction},
  \bibinfo{journal}{Measurement Science and Technology}
  \bibinfo{volume}{34}~(\bibinfo{number}{7})
  (\bibinfo{year}{2023}{\natexlab{b}}) \bibinfo{pages}{075017},
  \bibinfo{note}{\doi{https://doi.org/10.1088/1361-6501/accbdb}}.

\bibitem[{Yan et~al.(2022{\natexlab{b}})Yan, Fu, Lu, Li, Shen, and Wang}]{h78}
\bibinfo{author}{T.~Yan}, \bibinfo{author}{Y.~Fu}, \bibinfo{author}{M.~Lu},
  \bibinfo{author}{Z.~Li}, \bibinfo{author}{C.~Shen},
  \bibinfo{author}{D.~Wang}, \bibinfo{title}{Integration of a Novel
  Knowledge-Guided Loss Function With an Architecturally Explainable Network
  for Machine Degradation Modeling}, \bibinfo{journal}{IEEE Transactions on
  Instrumentation and Measurement} \bibinfo{volume}{71}
  (\bibinfo{year}{2022}{\natexlab{b}}) \bibinfo{pages}{1--12},
  \bibinfo{note}{\doi{https://doi.org/10.1109/TIM.2022.3193196}}.

\bibitem[{Russell and Wang(2022)}]{h79}
\bibinfo{author}{M.~Russell}, \bibinfo{author}{P.~Wang},
  \bibinfo{title}{Physics-informed deep learning for signal compression and
  reconstruction of big data in industrial condition monitoring},
  \bibinfo{journal}{Mechanical Systems and Signal Processing}
  \bibinfo{volume}{168} (\bibinfo{year}{2022}) \bibinfo{pages}{108709},
  \bibinfo{note}{\doi{https://doi.org/10.1016/j.ymssp.2021.108709}}.

\bibitem[{Chen et~al.(2022)Chen, Ma, Zhao, Zhai, and Mao}]{h80}
\bibinfo{author}{X.~Chen}, \bibinfo{author}{M.~Ma}, \bibinfo{author}{Z.~Zhao},
  \bibinfo{author}{Z.~Zhai}, \bibinfo{author}{Z.~Mao},
  \bibinfo{title}{Physics-Informed Deep Neural Network for Bearing Prognosis
  with Multisensory Signals}, \bibinfo{journal}{Journal of Dynamics, Monitoring
  and Diagnostics} \bibinfo{volume}{1}~(\bibinfo{number}{4})
  (\bibinfo{year}{2022}) \bibinfo{pages}{200–207},
  \bibinfo{note}{\doi{https://doi.org/10.37965/jdmd.2022.54}}.

\bibitem[{Freeman et~al.(2022)Freeman, Tang, Huang, and VanZwieten}]{h81}
\bibinfo{author}{B.~Freeman}, \bibinfo{author}{Y.~Tang},
  \bibinfo{author}{Y.~Huang}, \bibinfo{author}{J.~VanZwieten},
  \bibinfo{title}{Physics-informed turbulence intensity infusion: A new hybrid
  approach for marine current turbine rotor blade fault detection},
  \bibinfo{journal}{Ocean Engineering} \bibinfo{volume}{254}
  (\bibinfo{year}{2022}) \bibinfo{pages}{111299},
  \bibinfo{note}{\doi{https://doi.org/10.1016/j.oceaneng.2022.111299}}.

\bibitem[{Xu and Noh(2021)}]{h82}
\bibinfo{author}{S.~Xu}, \bibinfo{author}{H.~Y. Noh}, \bibinfo{title}{PhyMDAN:
  Physics-informed knowledge transfer between buildings for seismic damage
  diagnosis through adversarial learning}, \bibinfo{journal}{Mechanical Systems
  and Signal Processing} \bibinfo{volume}{151} (\bibinfo{year}{2021})
  \bibinfo{pages}{107374},
  \bibinfo{note}{\doi{https://doi.org/10.1016/j.ymssp.2020.107374}}.

\bibitem[{Zhang and Sun(2021)}]{h83}
\bibinfo{author}{Z.~Zhang}, \bibinfo{author}{C.~Sun},
  \bibinfo{title}{Structural damage identification via physics-guided machine
  learning: a methodology integrating pattern recognition with finite element
  model updating}, \bibinfo{journal}{Structural Health Monitoring}
  \bibinfo{volume}{20}~(\bibinfo{number}{4}) (\bibinfo{year}{2021})
  \bibinfo{pages}{1675--1688},
  \bibinfo{note}{\doi{https://doi.org/10.1177/1475921720927488}}.

\bibitem[{Jablonski and Dziedziech(2022)}]{h39}
\bibinfo{author}{A.~Jablonski}, \bibinfo{author}{K.~Dziedziech},
  \bibinfo{title}{Intelligent spectrogram--A tool for analysis of complex
  non-stationary signals}, \bibinfo{journal}{Mechanical Systems and Signal
  Processing} \bibinfo{volume}{167} (\bibinfo{year}{2022})
  \bibinfo{pages}{108554},
  \bibinfo{note}{\doi{https://doi.org/10.1016/j.ymssp.2021.108554}}.

\bibitem[{R{\'e}nyi(1961)}]{h41}
\bibinfo{author}{A.~R{\'e}nyi}, \bibinfo{title}{On measures of entropy and
  information}, in: \bibinfo{booktitle}{Proceedings of the Fourth Berkeley
  Symposium on Mathematical Statistics and Probability, Volume 1: Contributions
  to the Theory of Statistics}, vol.~\bibinfo{volume}{4},
  \bibinfo{pages}{547--562},
  \bibinfo{note}{\doi{https://static.renyi.hu/renyi\_cikkek/1961\_on\_measures\_of\_entropy\_and\_information.pdf}},
  \bibinfo{year}{1961}.

\bibitem[{Marx and Gryllias(2023)}]{h40}
\bibinfo{author}{D.~Marx}, \bibinfo{author}{K.~Gryllias},
  \bibinfo{title}{Differentiable Short-Time Fourier Transform Window Length
  Selection Driven by Cyclo-Stationarity}, in: \bibinfo{booktitle}{Annual
  Conference of the PHM Society}, vol.~\bibinfo{volume}{15},
  \bibinfo{note}{\doi{https://doi.org/10.36001/phmconf.2023.v15i1.3566}},
  \bibinfo{year}{2023}.

\bibitem[{Leiber et~al.(2023)Leiber, Marnissi, Barrau, and Badaoui}]{h37}
\bibinfo{author}{M.~Leiber}, \bibinfo{author}{Y.~Marnissi},
  \bibinfo{author}{A.~Barrau}, \bibinfo{author}{M.~E. Badaoui},
  \bibinfo{title}{Differentiable Adaptive Short-Time Fourier Transform with
  Respect to the Window Length}, in: \bibinfo{booktitle}{{IEEE} International
  Conference on Acoustics, Speech and Signal Processing {ICASSP} 2023, Rhodes
  Island, Greece, June 4-10, 2023}, \bibinfo{pages}{1--5},
  \bibinfo{note}{\doi{https://doi.org/10.1109/ICASSP49357.2023.10095245}},
  \bibinfo{year}{2023}.

\bibitem[{Huang and Baddour(2018)}]{h43}
\bibinfo{author}{H.~Huang}, \bibinfo{author}{N.~Baddour},
  \bibinfo{title}{Bearing vibration data collected under time-varying
  rotational speed conditions}, \bibinfo{journal}{Data in brief}
  \bibinfo{volume}{21} (\bibinfo{year}{2018}) \bibinfo{pages}{1745--1749},
  \bibinfo{note}{\doi{https://doi.org/10.1016/j.dib.2018.11.019}}.

\bibitem[{Tan and Le(2021)}]{h49}
\bibinfo{author}{M.~Tan}, \bibinfo{author}{Q.~V. Le},
  \bibinfo{title}{EfficientNetV2: Smaller Models and Faster Training}, in:
  \bibinfo{booktitle}{Proceedings of the 38th International Conference on
  Machine Learning, {ICML} 2021, 18-24 July 2021, Virtual Event}, vol.
  \bibinfo{volume}{139} of \emph{\bibinfo{series}{Proceedings of Machine
  Learning Research}}, \bibinfo{publisher}{{PMLR}},
  \bibinfo{pages}{10096--10106},
  \bibinfo{note}{\doi{http://proceedings.mlr.press/v139/tan21a.html}},
  \bibinfo{year}{2021}.

\bibitem[{He et~al.(2016)He, Zhang, Ren, and Sun}]{h50}
\bibinfo{author}{K.~He}, \bibinfo{author}{X.~Zhang}, \bibinfo{author}{S.~Ren},
  \bibinfo{author}{J.~Sun}, \bibinfo{title}{Deep residual learning for image
  recognition}, in: \bibinfo{booktitle}{Proceedings of the IEEE conference on
  computer vision and pattern recognition}, \bibinfo{pages}{770--778},
  \bibinfo{note}{\doi{https://doi.org/10.1109/CVPR.2016.90}},
  \bibinfo{year}{2016}.

\bibitem[{Huang et~al.(2017)Huang, Liu, van~der Maaten, and Weinberger}]{h51}
\bibinfo{author}{G.~Huang}, \bibinfo{author}{Z.~Liu},
  \bibinfo{author}{L.~van~der Maaten}, \bibinfo{author}{K.~Q. Weinberger},
  \bibinfo{title}{Densely Connected Convolutional Networks}, in:
  \bibinfo{booktitle}{2017 {IEEE} Conference on Computer Vision and Pattern
  Recognition, {CVPR} 2017, Honolulu, HI, USA, July 21-26, 2017},
  \bibinfo{pages}{2261--2269},
  \bibinfo{note}{\doi{https://doi.org/10.1109/CVPR.2017.243}},
  \bibinfo{year}{2017}.

\bibitem[{Zhang et~al.(2017)Zhang, Peng, Li, Chen, and Zhang}]{h52}
\bibinfo{author}{W.~Zhang}, \bibinfo{author}{G.~Peng}, \bibinfo{author}{C.~Li},
  \bibinfo{author}{Y.~Chen}, \bibinfo{author}{Z.~Zhang}, \bibinfo{title}{A new
  deep learning model for fault diagnosis with good anti-noise and domain
  adaptation ability on raw vibration signals}, \bibinfo{journal}{Sensors}
  \bibinfo{volume}{17}~(\bibinfo{number}{2}) (\bibinfo{year}{2017})
  \bibinfo{pages}{425},
  \bibinfo{note}{\doi{https://doi.org/10.3390/s17020425}}.

\bibitem[{Zhao et~al.(2020)Zhao, Zhong, Fu, Tang, and Pecht}]{h53}
\bibinfo{author}{M.~Zhao}, \bibinfo{author}{S.~Zhong}, \bibinfo{author}{X.~Fu},
  \bibinfo{author}{B.~Tang}, \bibinfo{author}{M.~Pecht}, \bibinfo{title}{Deep
  Residual Shrinkage Networks for Fault Diagnosis}, \bibinfo{journal}{IEEE
  Transactions on Industrial Informatics}
  \bibinfo{volume}{16}~(\bibinfo{number}{7}) (\bibinfo{year}{2020})
  \bibinfo{pages}{4681--4690},
  \bibinfo{note}{\doi{https://doi.org/10.1109/TII.2019.2943898}}.

\bibitem[{Xiao et~al.(2022)Xiao, Shao, Han, Huo, and Wan}]{h57}
\bibinfo{author}{Y.~Xiao}, \bibinfo{author}{H.~Shao}, \bibinfo{author}{S.~Han},
  \bibinfo{author}{Z.~Huo}, \bibinfo{author}{J.~Wan}, \bibinfo{title}{Novel
  Joint Transfer Network for Unsupervised Bearing Fault Diagnosis From
  Simulation Domain to Experimental Domain}, \bibinfo{journal}{IEEE/ASME on
  Mechatronics} \bibinfo{volume}{27}~(\bibinfo{number}{6})
  (\bibinfo{year}{2022}) \bibinfo{pages}{5254--5263},
  \bibinfo{note}{\doi{https://doi.org/10.1109/TMECH.2022.3177174}}.

\bibitem[{Qian et~al.(2023{\natexlab{a}})Qian, Wang, Zhang, and Qin}]{h55}
\bibinfo{author}{Q.~Qian}, \bibinfo{author}{Y.~Wang},
  \bibinfo{author}{T.~Zhang}, \bibinfo{author}{Y.~Qin}, \bibinfo{title}{Maximum
  mean square discrepancy: a new discrepancy representation metric for
  mechanical fault transfer diagnosis}, \bibinfo{journal}{Knowledge-Based
  Systems} \bibinfo{volume}{276} (\bibinfo{year}{2023}{\natexlab{a}})
  \bibinfo{pages}{110748},
  \bibinfo{note}{\doi{https://doi.org/10.1016/j.knosys.2023.110748}}.

\bibitem[{Qian et~al.(2023{\natexlab{b}})Qian, Qin, Luo, Wang, and Wu}]{h56}
\bibinfo{author}{Q.~Qian}, \bibinfo{author}{Y.~Qin}, \bibinfo{author}{J.~Luo},
  \bibinfo{author}{Y.~Wang}, \bibinfo{author}{F.~Wu}, \bibinfo{title}{Deep
  discriminative transfer learning network for cross-machine fault diagnosis},
  \bibinfo{journal}{Mechanical Systems and Signal Processing}
  \bibinfo{volume}{186} (\bibinfo{year}{2023}{\natexlab{b}})
  \bibinfo{pages}{109884},
  \bibinfo{note}{\doi{https://doi.org/10.1016/j.ymssp.2022.109884}}.

\bibitem[{Ravanelli and Bengio(2018)}]{h58}
\bibinfo{author}{M.~Ravanelli}, \bibinfo{author}{Y.~Bengio},
  \bibinfo{title}{Speaker Recognition from Raw Waveform with SincNet}, in:
  \bibinfo{booktitle}{2018 IEEE Spoken Language Technology Workshop (SLT)},
  \bibinfo{pages}{1021--1028},
  \bibinfo{note}{\doi{https://doi.org/10.1109/SLT.2018.8639585}},
  \bibinfo{year}{2018}.

\bibitem[{Liu et~al.(2024)Liu, Ding, Wu, He, and Shao}]{h831}
\bibinfo{author}{R.~Liu}, \bibinfo{author}{X.~Ding}, \bibinfo{author}{Q.~Wu},
  \bibinfo{author}{Q.~He}, \bibinfo{author}{Y.~Shao}, \bibinfo{title}{An
  Interpretable Multiplication-Convolution Network for Equipment Intelligent
  Edge Diagnosis}, \bibinfo{journal}{IEEE Transactions on Systems, Man, and
  Cybernetics: Systems}  (\bibinfo{year}{2024})
  \bibinfo{pages}{1--12}\bibinfo{note}{\doi{https://doi.org/10.1109/TSMC.2023.3346398}}.

\bibitem[{Liu et~al.(2022)Liu, Ma, Han, Shi, Qin, and Hu}]{h59}
\bibinfo{author}{C.~Liu}, \bibinfo{author}{X.~Ma}, \bibinfo{author}{T.~Han},
  \bibinfo{author}{X.~Shi}, \bibinfo{author}{C.~Qin}, \bibinfo{author}{S.~Hu},
  \bibinfo{title}{NTScatNet: An interpretable convolutional neural network for
  domain generalization diagnosis across different transmission paths},
  \bibinfo{journal}{Measurement} \bibinfo{volume}{204} (\bibinfo{year}{2022})
  \bibinfo{pages}{112041},
  \bibinfo{note}{\doi{https://doi.org/10.1016/j.measurement.2022.112041}}.

\bibitem[{Shang et~al.(2023)Shang, Zhao, and Yan}]{h60}
\bibinfo{author}{Z.~Shang}, \bibinfo{author}{Z.~Zhao},
  \bibinfo{author}{R.~Yan}, \bibinfo{title}{Denoising Fault-Aware Wavelet
  Network: A Signal Processing Informed Neural Network for Fault Diagnosis},
  \bibinfo{journal}{Chinese Journal of Mechanical Engineering}
  \bibinfo{volume}{36}~(\bibinfo{number}{1}) (\bibinfo{year}{2023})
  \bibinfo{pages}{9},
  \bibinfo{note}{\doi{https://doi.org/10.1186/s10033-023-00838-0}}.

\bibitem[{Jia et~al.(2023)Jia, Chow, and Yuan}]{h61}
\bibinfo{author}{L.~Jia}, \bibinfo{author}{T.~W. Chow},
  \bibinfo{author}{Y.~Yuan}, \bibinfo{title}{GTFE-Net: A Gramian Time Frequency
  Enhancement CNN for bearing fault diagnosis}, \bibinfo{journal}{Engineering
  Applications of Artificial Intelligence} \bibinfo{volume}{119}
  (\bibinfo{year}{2023}) \bibinfo{pages}{105794},
  \bibinfo{note}{\doi{https://doi.org/10.1016/j.engappai.2022.105794}}.

\bibitem[{Alekseev and Bobe(2019)}]{h62}
\bibinfo{author}{A.~Alekseev}, \bibinfo{author}{A.~Bobe},
  \bibinfo{title}{GaborNet: Gabor filters with learnable parameters in deep
  convolutional neural network}, in: \bibinfo{booktitle}{2019 International
  Conference on Engineering and Telecommunication (EnT)},
  \bibinfo{pages}{1--4},
  \bibinfo{note}{\doi{https://doi.org/10.1109/EnT47717.2019.9030571}},
  \bibinfo{year}{2019}.

\bibitem[{He et~al.(2024)He, Shi, Liu, and Li}]{h67}
\bibinfo{author}{C.~He}, \bibinfo{author}{H.~Shi}, \bibinfo{author}{X.~Liu},
  \bibinfo{author}{J.~Li}, \bibinfo{title}{Interpretable physics-informed
  domain adaptation paradigm for cross-machine transfer diagnosis},
  \bibinfo{journal}{Knowledge-Based Systems} \bibinfo{volume}{288}
  (\bibinfo{year}{2024}) \bibinfo{pages}{111499},
  \bibinfo{note}{\doi{https://doi.org/10.1016/j.knosys.2024.111499}}.

\bibitem[{Yue et~al.(2023)Yue, Li, Chen, Huang, and Li}]{h63}
\bibinfo{author}{K.~Yue}, \bibinfo{author}{J.~Li}, \bibinfo{author}{J.~Chen},
  \bibinfo{author}{R.~Huang}, \bibinfo{author}{W.~Li},
  \bibinfo{title}{Multiscale Wavelet Prototypical Network for Cross-Component
  Few-Shot Intelligent Fault Diagnosis}, \bibinfo{journal}{IEEE Transactions on
  Instrumentation and Measurement} \bibinfo{volume}{72} (\bibinfo{year}{2023})
  \bibinfo{pages}{1--11},
  \bibinfo{note}{\doi{https://doi.org/10.1109/TIM.2022.3230480}}.

\bibitem[{Chen et~al.(2024)Chen, Dong, Tu, Wang, Cheng, Zhao, and Peng}]{h64}
\bibinfo{author}{Q.~Chen}, \bibinfo{author}{X.~Dong}, \bibinfo{author}{G.~Tu},
  \bibinfo{author}{D.~Wang}, \bibinfo{author}{C.~Cheng},
  \bibinfo{author}{B.~Zhao}, \bibinfo{author}{Z.~Peng}, \bibinfo{title}{TFN: An
  interpretable neural network with time-frequency transform embedded for
  intelligent fault diagnosis}, \bibinfo{journal}{Mechanical Systems and Signal
  Processing} \bibinfo{volume}{207} (\bibinfo{year}{2024})
  \bibinfo{pages}{110952},
  \bibinfo{note}{\doi{https://doi.org/10.1016/j.ymssp.2023.110952}}.

\bibitem[{Shi et~al.(2021)Shi, Chen, Zi, and Zhou}]{h65}
\bibinfo{author}{Z.~Shi}, \bibinfo{author}{J.~Chen}, \bibinfo{author}{Y.~Zi},
  \bibinfo{author}{Z.~Zhou}, \bibinfo{title}{A Novel Multitask Adversarial
  Network via Redundant Lifting for Multicomponent Intelligent Fault Detection
  Under Sharp Speed Variation}, \bibinfo{journal}{IEEE Transactions on
  Instrumentation and Measurement} \bibinfo{volume}{70} (\bibinfo{year}{2021})
  \bibinfo{pages}{1--10},
  \bibinfo{note}{\doi{https://doi.org/10.1109/TIM.2021.3055821}}.

\end{thebibliography}

\end{document}